\documentclass[10pt,a4paper]{article}

\usepackage{amssymb}
\usepackage{graphicx}
\usepackage{enumerate}
\usepackage{adjustbox}
\usepackage{soul}
\usepackage{relsize}
\usepackage{mathtools}
\usepackage{xstring}
\usepackage{cases}
\usepackage{multirow}
\usepackage{multicol}
\usepackage{subfloat}
\usepackage{subfig}
\usepackage{caption}
\usepackage{marvosym}
\usepackage{url}
\usepackage{booktabs}
\usepackage[textwidth=14cm]{geometry}

\DeclareMathAlphabet\mathbfcal{OMS}{cmsy}{b}{n}

  {\arraycolsep 0pt \begin{array}{l}}%
  {\end{array}}

\usepackage{calc}
\newlength{\depthofsumsign}
\newcommand{\predArg}[1]{
\IfSubStr{#1}{(}{
  \mathtt{\StrBefore{#1}{(}}\mathit{(\StrBehind{#1}{(}}}
  {
  \mathit{#1}
  }
}

\def\folhappensAt{\mathtt{HappensAt}}
\def\folholdsAt{\mathtt{HoldsAt}}
\def\folinitiatedAt{\mathtt{InitiatedAt}}
\def\folterminatedAt{\mathtt{TerminatedAt}}

\newcommand{\pHappensAt}[2]{\folhappensAt(\predArg{#1},\, #2)}
\newcommand{\pInitiatedAt}[2]{\folinitiatedAt(\predArg{#1},\, #2)}
\newcommand{\pTerminatedAt}[2]{\folterminatedAt(\predArg{#1},\, #2)}

\def\PEC{\mathtt{MLN}{\!-\!}\mathtt{EC}}
\def\MLNEC{$\PEC$}
\def\OSLa{$\mathtt{OSL}\alpha$}

\begin{document}

\title{An Integrated and Scalable Platform for Proactive Event-Driven Traffic Management}

\author{Alain Kibangou \and Alexander Artikis \and Evangelos Michelioudakis \and Georgios Paliouras \and Marius Schmitt \and John Lygeros \and Chris Baber \and Natan Morar \and Fabiana Fournier \and Inna Skarbovsky}

\date{}

\maketitle

\begin{abstract}
Traffic on freeways can be managed by means of ramp meters from Road Traffic Control rooms. Human operators cannot efficiently manage a network of ramp meters. To support them, we present an intelligent platform for traffic management which includes a new ramp metering coordination scheme in the decision making module, an efficient dashboard for interacting with human operators, machine learning tools for learning event definitions and Complex Event Processing tools able to deal with uncertainties inherent to the traffic use case. Unlike the usual approach, the devised event-driven platform is able to predict a congestion up to $4$ minutes before it really happens. Proactive decision making can then be established leading to significant improvement of traffic conditions.
\end{abstract}

\section{Introduction}
Congestion can be defined as a situation when traffic is moving at speed below the designed capacity of a roadway \cite{Downs2004} or as a state of traffic flow on a transportation facility characterized by high densities and low speeds, relative to some chosen reference state \cite{BovySalomon2002}. It results of various root causes (e.g. traffic incidents, work zones, weather, special events, physical bottlenecks), often interacting with one another \cite{Downs2004} and induces excess delays, reduced safety, and increased environmental pollution due to stop-and-go behaviour. One approach to tackle congestion could be to increase the capacity of the traffic infrastructure by constructing new roads. This approach is very costly and it is often not possible due to societal constraints as citizens are more and more aware of environment protection. The solution is then to control traffic in order to avoid, reduce or at least postpone congestion. To do so, most cities in the world have taken important decisions to invest in road sensor capabilities to get measurements of traffic parameters and to build modern Road Traffic Control rooms where traffic operators monitor the traffic situation based on video images and measurements from loop detectors and wireless magnetic sensors, for instance \cite{Kojima1999}. Actions to control traffic are two-fold: manage ramp metering and/or change speed limits according to the current traffic status \cite{pisarski:hal-00727783}. We will focus on ramp metering which is the most common regulation policy.

\par As stated above, existing traffic management platforms are mainly based on human operators and noisy data arriving from various sensors \cite{Kojima1999}. They are based on the paradigm \textit{sense-respond}. In contrast, here, we consider a \textit{sense-recognise-forecast-decide-act-explain} paradigm where decisions are triggered by forecasting events, whether they correspond to problems or opportunities, instead of reacting to them once they happen. We present \textit{SPEEDD (Scalable Proactive Event-Driven Decision Making)},  an integrated platform for proactive event-driven decision-making and demonstrate its capabilities to be resilient to the inherent uncertainty of the sensor readings, which include incomplete data streams, erroneous data and imprecise definitions of the events that need to be detected and/or forecasted. The following steps are to be considered. First, data are continuously acquired from various types of sensor and fused in order to recognise, in real-time, events of special significance. Second, the recognised events are correlated with historical information to forecast congestion that may take place in the near future. Third, both forecasted and recognised events are leveraged for real-time operational decision-making. Fourth, visual analytics \cite{ThomasCook2005} prioritise and explain possible proactive actions, enabling human operators to reach and execute the correct decision. The novelty of the proposed platform lies in the difficult task of on-the-fly, low-latency processing of large, geographically distributed, noisy event streams and historical data, for recognising and forecasting congestion, making decisions to reduce the impact of the congestion, and explaining the decisions to human operators in order to facilitate correct decision execution.

\par Operators in Road Traffic Control rooms have to monitor several on-ramps and actions are in general restricted to the identified bottleneck. In absence of coordination of all the ramp meters, the on-ramp immediately upstream of a bottleneck will solely attempt to prevent a congestion forming at the bottleneck. This local control often results in a quick growth of the queue length on the ramp. Then, to avoid an unacceptable spill of the congestion in the nearby urban area, the metering action needs to be limited, resulting in a congestion starting at the bottleneck and propagating upstream. By contrast, coordination between the ramps allows to distribute the control burden onto multiple ramps, thereby preventing ramp overflow without causing a congestion on the mainline. The main challenge is to determine when it is necessary to use on-ramps to hold vehicles back in the queue and reduce traffic demand from a downstream bottleneck. Such an action necessarily has to happen in a proactive way, since any effects of ramp metering travel downstream at most with the free-flow speed. SPEEDD supports a hierarchical coordination scheme with predictions made by means of complex event processing tools. The hierarchical coordination scheme decomposes the controller into distributed, local feedback loops, and a high-level coordination scheme based on events. Unlike existing approaches \cite{papamichail2010heuristic,papamichail2010coordinated}, it is based on the optimality analysis of decentralized ramp metering carried out in \cite{schmitt2016sufficient}. 

\par The remainder of this paper is organized as follows: in Section \ref{Chap_scenario} we describe the scenario under study. Then, following the \textit{sense-recognise-forecast-decide-act-explain} paradigm we describe the methods for event recognition and forecasting in Section \ref{Chap_Detection}, decision making in Section \ref{Chap_Decision}, and dashboard design in Section \ref{Chap_Interface}. Based on the developed method, we describe the integrated prototype in Section \ref{Chap_Architecture} and evaluate its performance in Section \ref{Chap_Evaluation}. Finally in Section \ref{Chap_Conclusions} we propose directions for future work and conclude.

\section{Scenario description}\label{Chap_scenario}
\begin{figure}[hbtp]
\centerline{\includegraphics[scale = 0.46]{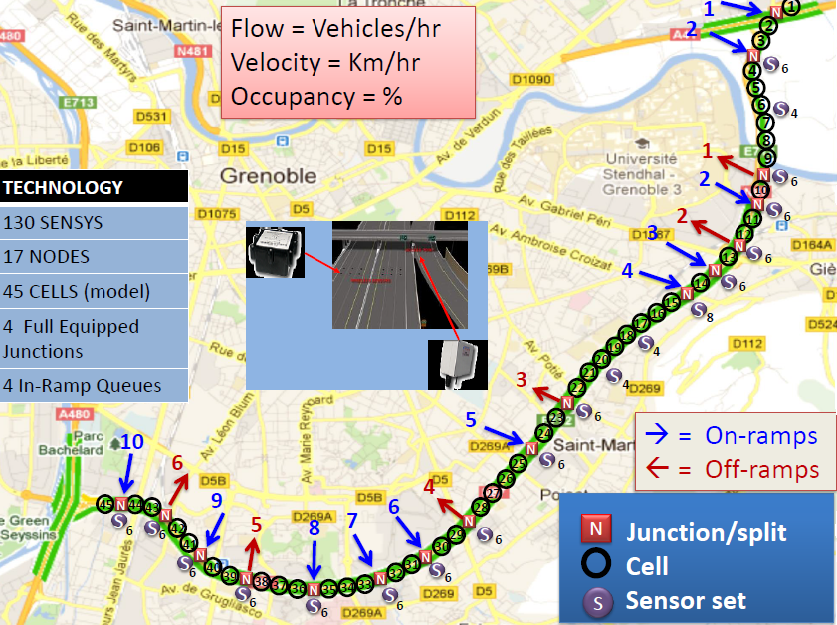}}
\caption{Grenoble South Ring Network: the road is divided in 45 cells numbered from east to west;  Cells equipped with sensors have an $S$ symbol; nodes, marked with $N$, are constituted by on-ramps (blue arrow) or off-ramps (red arrow).}
\label{fig_GTL}
\end{figure}

We consider the Grenoble South Ring road, in France, as the case study. This freeway links the city of Grenoble from the north-east to the south-west. In addition to sustaining local traffic, it has a major role since it connects two highways: the A480, which goes from Paris and Lyon to Marseille, and the A41, which goes from Grenoble to Switzerland. Moreover, the mountains surrounding Grenoble prevent the development of new roads, and also have a negative impact on pollution dispersion, making the problem of traffic regulation on this road even more crucial. From the Road traffic Control room of DIR CE\footnote{Direction Interd\'{e}partementale des Routes Centre-Est}, operators can monitor traffic on the road network, including the South Ring road, though hundreds of CCTV cameras, verbal reports (primarily from traffic operators on the road but also from police or other emergency service personnel), emails and other text messages, and data from sensors placed on the road. They can also effect traffic though variable message signs (VMS) and soon they will be involved in management of ramp metering, which is still under deployment.
DIR CE is also a partner of the GTL (Grenoble Traffic Lab), which offers a dense network of wireless magnetic sensors (see Fig. \ref{fig_GTL} and \cite{canudasdewit:hal-01059126} for a full description of the sensing platform). GTL also provides a microscopic calibrated simulator of the Grenoble south ring where the dynamics of each each vehicle in the road are simulated. This simulator has been developed using the AIMSUN platform\footnote{\url{http://www.aimsun.com/}}. It gives the opportunity to test the entire system in closed loop, from sensing to actuation; which is not possible with the actual freeway. In addition, synthetic data produced by the simulator have annotations that can serve as baseline in order to test the effectiveness and efficiency of the developed system.
\par Our objective is to detect congestions a few minutes before they happen. So, proactive suggestions to traffic operators can be provided, or automatic actions can be carried out, to alleviate the forecasted congestion. The following sections describe the methods allowing us to reach this objective.

\section{Event-driven Congestion detection and forecasting}\label{Chap_Detection}
The detection of congestions is based on information received from the sensors. Special behaviors of variables that describe the system such as speed, density, occupancy allow to infer the existence of congestion. In what follows, we adopt a Complex Event Processing (CEP) approach, sometimes called event stream processing, which is a method that combines data from multiple sources for tracking and analyzing (processing) streams of information (data) to infer events or patterns that suggest more complicated circumstances. The goal of complex event processing is to identify meaningful events (such as opportunities or threats) and respond to them as quickly as possible \cite{Luckham:2012}. 
In general, there exist two methods to define the rule patterns for a CEP application: machine learning and domain experts. In the first case, the patterns are learnt automatically by a computer program, while in the second, they are given by an external entity; usually a subject expert matter specialized in the domain. It is also possible to combine these two methods. Historical data used at design time contain raw events reported during the observed period along with annotations provided by domain experts. These annotations mark important situations that have been observed in the past and should be detected automatically in the future. Due to the dynamic nature of the proactive traffic management application, the knowledge base of event pattern definitions may require to be refined or enhanced with new ones. 
\subsection{Machine learning for event definitions}\label{machine_learning}
\begin{figure}[h]
\centering
\includegraphics[scale=0.6]{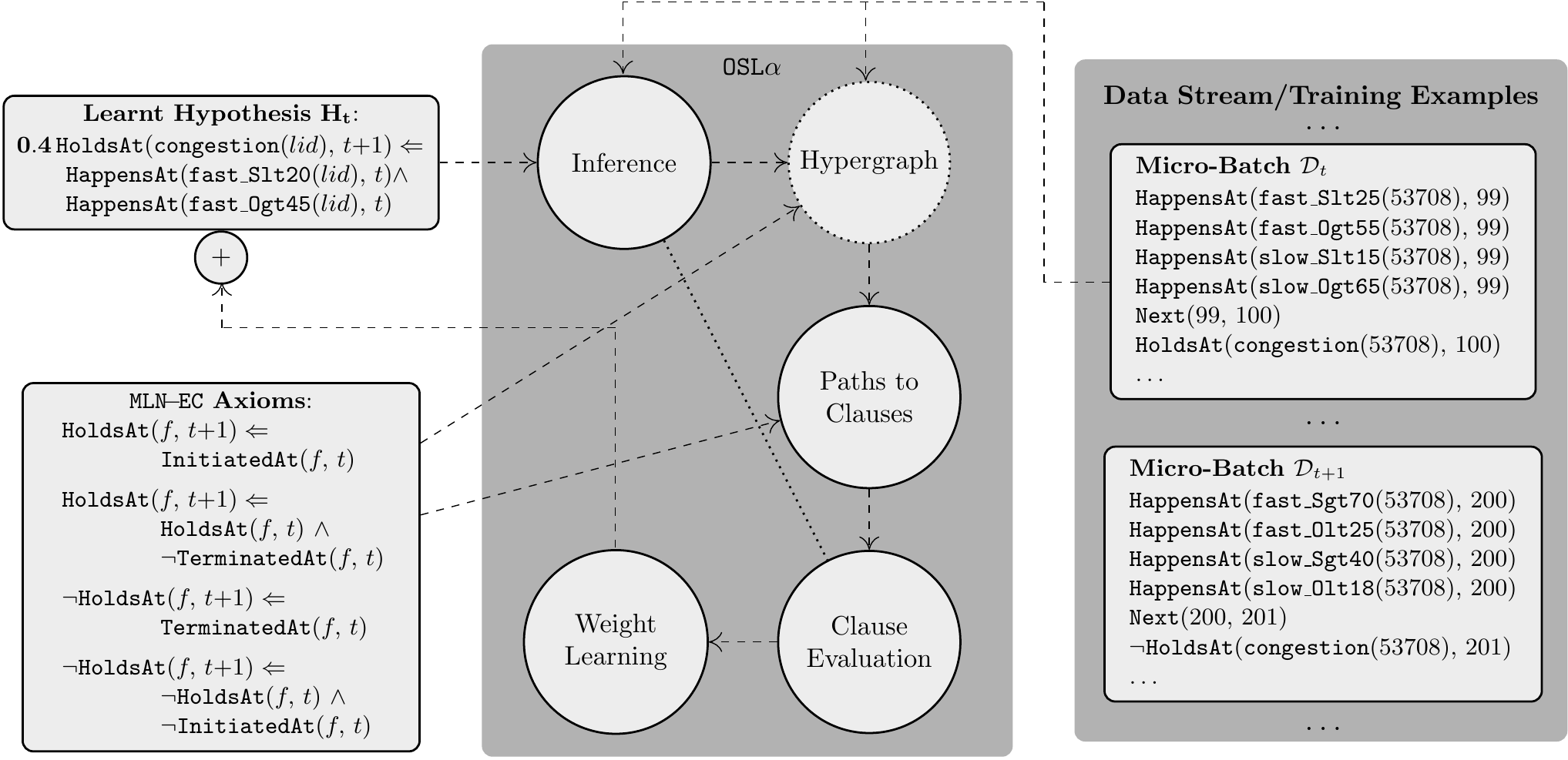}
\caption{The procedure of \OSLa.}
\label{fig:osla}
\end{figure}

In order to effectively learn definitions for traffic congestion using sensor data, we have developed \OSLa~\cite{vagmcs2016osla}, an online structure learner for Markov Logic Networks (MLNs) \cite{domingos2006markov}. \OSLa\ extends the procedure of OSL \cite{huynh2011osl} by exploiting a given background knowledge to effectively constrain the space of possible structures during learning. The space is constrained subject to characteristics imposed by the rules governing a specific task, herein stated as axioms. As a background knowledge we make use of \MLNEC~\cite{anskarl-TOCL15}, a probabilistic variant of the Event Calculus \cite{kowalksi1986EC,mueller2008} for event recognition. Fig. \ref{fig:osla} presents the components of \OSLa.

The background knowledge consists of the \MLNEC\ axioms (i.e., domain-independent rules) and an already known (possibly empty) hypothesis (i.e., set of clauses). Each axiom contains \textit{query predicates} $\folholdsAt \in \mathcal{Q}$ that consist of the supervision and \textit{template predicates} $\folinitiatedAt$, $\folterminatedAt \in \mathcal{P}$ that specify the conditions under which a complex event starts and stops being recognized. The latter form the target complex event definitions that we want to learn. \OSLa\ exploits these axioms in order to create mappings of supervision predicates into template predicates and search only for explanations of these template predicates. Upon doing so, \OSLa\ does not need to search over time sequences, instead only needs to find appropriate bodies over the current time-point for the following definite clauses:

\begin{align*}
& \pInitiatedAt{f}{t} \Leftarrow \mathrm{body}\\[2pt]
& \pTerminatedAt{f}{t} \Leftarrow \mathrm{body}
\end{align*}

At any step $t$ of the online procedure a training example (micro-batch) $\mathbfcal{D}_t$ arrives containing sensor readings, e.g. a fast lane in a highway has average speed less than $25$ km/hour and sensor occupancy greater than $55\%$. $\mathbfcal{D}_t$ is used together with the already learnt hypothesis to predict the truth values $y_t^P$ of the complex events of interest. This is achieved by (maximum a posteriori) MAP inference based on LP-relaxed Integer Linear Programming \cite{huynh2009max}. Given $\mathbfcal{D}_t$ \OSLa\ constructs a hypergraph that represents the space of possible structures as graph paths. Then for all incorrectly predicted complex events the hypergraph is searched, guided by \MLNEC\ axioms and path mode declarations \cite{huynh2011osl} using relational pathfinding \cite{richards1992RP} up to a predefined length, for definite clauses explaining these complex events. The paths discovered during the search correspond to conjunctions of true ground atoms and are generalized into first-order clauses by replacing constants in the conjunction with variables. Then, these conjunctions are used as a $\mathrm{body}$ to form definite clauses using as head the template predicate present in each path. The resulting set of formulas is converted into clausal normal form and evaluated. The weights of the retained clauses are then optimized by the AdaGrad online learner \cite{duchi2011AdaGrad}. Finally, the weighted clauses are appended to the hypothesis $\mathbfcal{H}_t$ and the procedure is repeated for the next training example $\mathbfcal{D}_{t+1}$.


\subsection{Event-driven approach to forecast congestions in real-time}
Event definitions learnt as described above are then used to forecast congestions in real-time by means of an event-driven application which can be defined by an event processing network (EPN) \cite{Etzion2010}. An EPN, a conceptual model describing the event processing flow execution, comprises a collection of event processing agents (EPAs), event producers, events, and consumers. The network describes the flow of events originating at event producers and flowing through various event processing agents to eventually reach event consumers. We resort to the IBM PROactive Technology ONline (PROTON\footnote{\url{https://github.com/ishkin/Proton/}}) as the CEP engine. In our scenario, the CEP component receives events emitted from the sensors (producers) every 15 seconds and based on predefined event rules, it alerts in case of a detection of a possible congestion. In this scenario, the input events are certain (a sensor reading event happens) but the derived event is not certain (e.g., the fact that we have 15 sensor readings in 5 minutes that show an increase in the density, doesn't necessarily imply there will be a traffic congestion for sure). In other words, the capability to forecast events requires the inclusion of uncertainty aspects. Proactive event-driven computing deals with the inherent uncertainty in the event inputs, in the output events, or in both (\cite{Artikis2012,Wasserkrug2012,Engel2012}).\newline
\par\noindent The EPN for the proposed system includes the following EPAs:
\begin{itemize}
\item \textit{Congestion} at a specific location:  it exists if the density in a specific location is above a certain given value ($density\_threshold1$) and the speed is below a certain given value ($speed\_threshold1$) for at least 15 input events within the time period of $5$ min or until a \textit{ClearCongestion} EPA occurs.
\item \textit{ClearCongestion} at a specific location: it occurs when a congestion is over, i.e. whenever the density is below a certain given value ($density\_threshold2$) and the speed is above a certain given value ($speed\_threshold2$) for at least 15 input events within a time window that is opened with either a \textit{Congestion} or a \textit{PredictedCongestion} events and is closed after $5$ min.
\item \textit{PredictedCongestion}: it occurs when a forecasted congestion is identified at a specific location. This event pattern or rule is probabilistic in the sense that the output or derived event has a certainty attribute value associated to it. It is of type TREND, meaning that it derives an event whenever a specific change (increasing or decreasing) over time of the density value in the input events is satisfied over a temporal window. This EPA emits a derived event if at least 5 input events show an increase in the density in a temporal window which is opened with the first input event that comes and is closed when either a \textit{Congestion} or \textit{ClearCongestion} is detected for the same location.
\item \textit{Calculations}: they concern calculations on sensor readings (such as averages). They are emitted to be consumed by the decision making module.
\end{itemize}
To cope with uncertainties inherent in the event rules, a Sigmoid function is used to calculate the confidence of the occurrence of a derived event. The idea is: whenever the number of events in the matching set of the TREND pattern in the \textit{PredictedCongestion} EPA is high enough, the certainty of the derived event is close to one.

\section{Decision making}\label{Chap_Decision}
\begin{figure}[hbtp]
	\centering
		\includegraphics[width=0.8\textwidth]{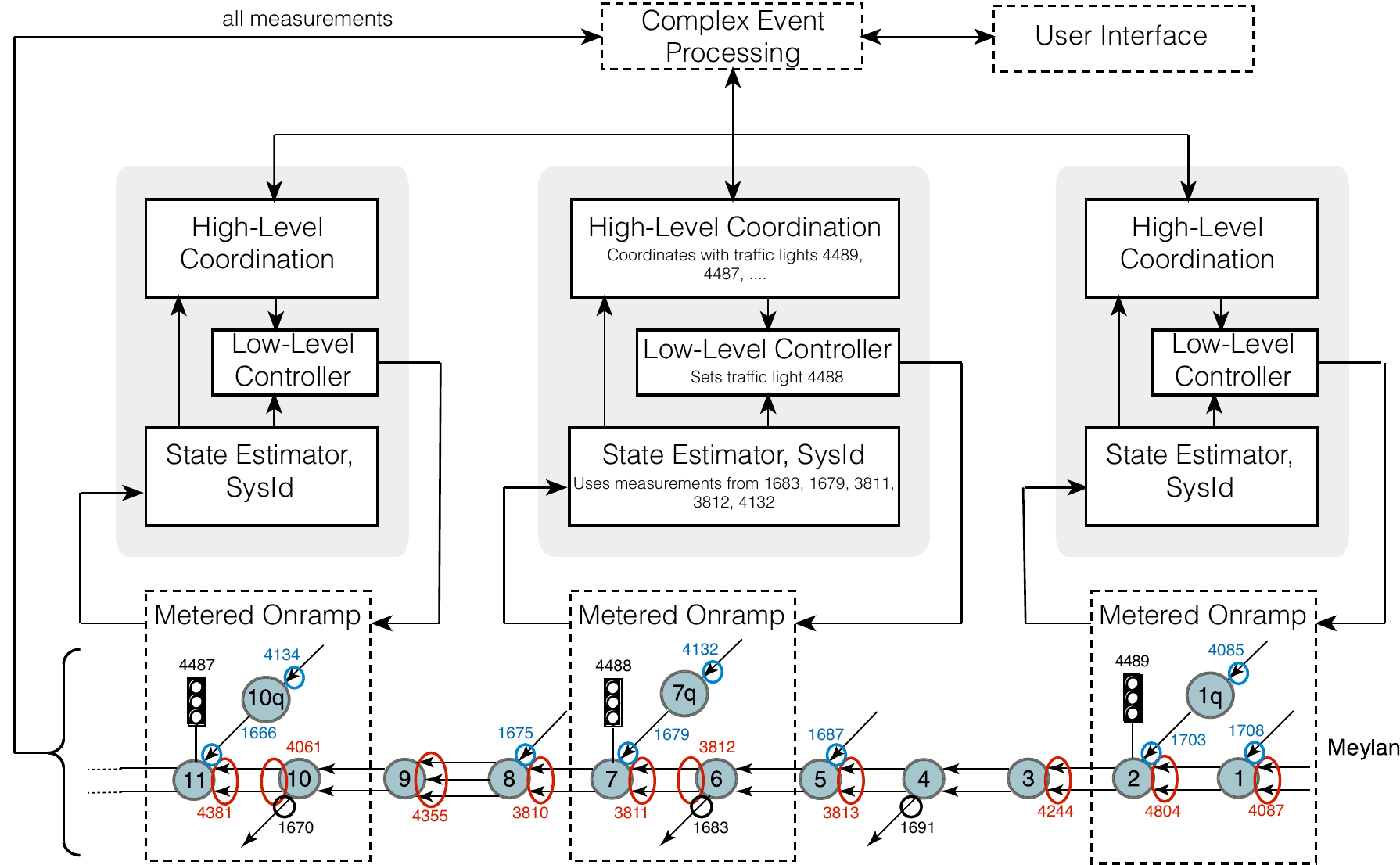}
	\caption{Hierarchical control approach for ramp metering.}
	\label{fig:control_structure}
\end{figure}

In the previous section, we have described how to derive smart rules to recognize and/or predict a congestion. Now, we will describe a smart way to manage or to avoid congestion. A cause of congestion is related to an excess of demand of using the road infrastructure. Such a demand can be managed by means of ramp metering. Given a set of equipped on-ramps, the objective is to regulate the entering flow in a smart way while avoiding congestion to spill back to the arterial network. We adopt a hierarchical approach, which decomposes the controller into distributed, local feedback loops, and a high-level coordination scheme based on events (see Fig. \ref{fig:control_structure}). Existing solutions are used for the local feedback laws but a new coordination scheme is proposed  to deal with non-monotonic effects in traffic dynamics. In other words, for monotonic traffic, local feedback laws are enough but the coordination scheme is necessary when non-monotonic effects occur.
The non-monotonic behavior makes the design of optimal controllers difficult, as non-convex problems arise. The main innovation in our approach lies in the usage of a model-free, data- and event-driven solution to the problem of ramp coordination.


\subsection{Data-driven System Identification} 
\label{sec:datadriven_fd}
\begin{figure}[hbtp]
\centering
\subfloat{\label{fig:fd_speedd_a}\includegraphics[width=0.4\textwidth]{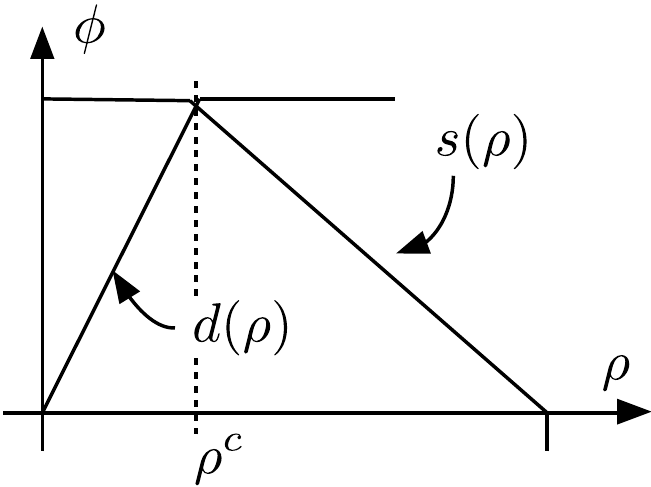}} \hspace{0.5cm}
\subfloat{\label{fig:fd_speedd_b}\includegraphics[width=0.4\textwidth]{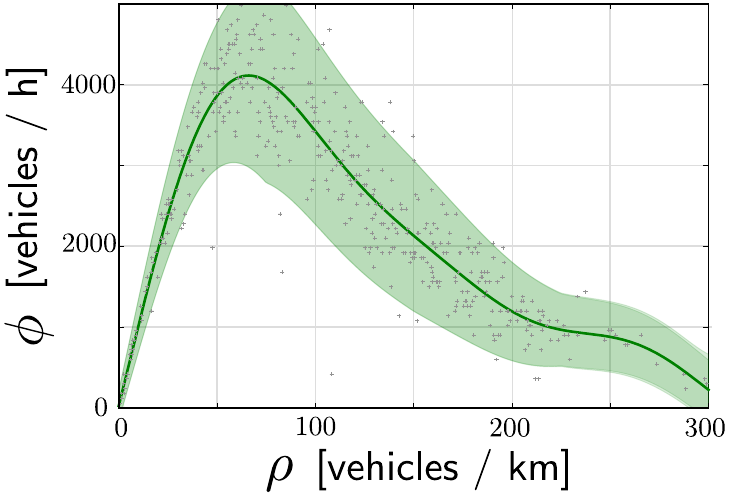}} \hspace{0.05cm}
\caption{Fundamental diagram: theoretical (left) and reconstructed from real data using GP regression with mean and 90\% confidence interval.}
\label{fig:fd_speedd}
\end{figure}

To understand how non-monotonic effects affect the ramp metering problem, it is necessary to briefly review road traffic dynamics. Freeway traffic conditions at some location can be described by the traffic density $\rho(t)$, measured in number of vehicles per kilometer, and the (mainline) traffic flow $\phi(t)$. First-order traffic model Lighthill-Whitham-Richards (LWR) \cite{lighthill1955kinematic} postulates a static flow-density relationship, which is called the \textit{fundamental diagram}. It is usual to associate a piecewise-affine fundamental diagram to the LWR model, a shape confirmed with real data but with significant levels of variance (see Fig. \ref{fig:fd_speedd}). This spread is most notable at and above the critical density (density value corresponding to the maximal flow) and it might partially be caused by a dependency of the flow on further variables.
\par
\begin{figure}[hbtp]
\centering
\includegraphics[width=0.45\textwidth]{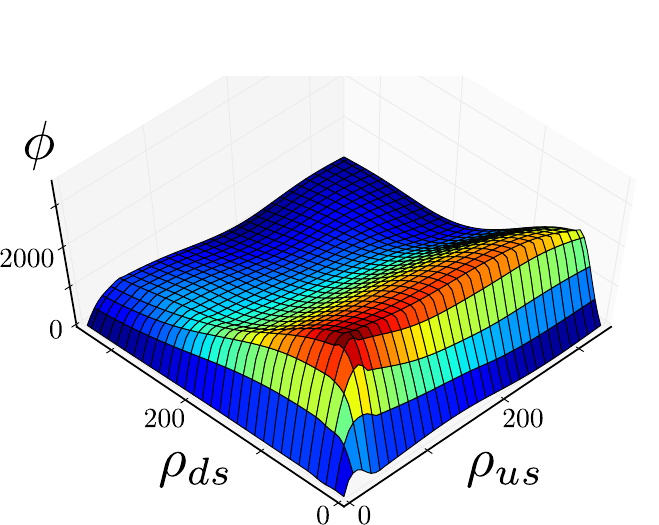}
\caption{GP regression for two-dimensional fundamental diagram where $\rho_{ds}$ (resp. $\rho_{us}$) stands for downstream density (resp. upstream density).}
\label{fig:fd_speedd_c}
\end{figure}

\noindent Considering a subdivision of the road in cells, the density of cell $k$ evolves as
\begin{equation}
\rho_k(t+1)=\rho_k(t)+\frac{T}{L_k}(\phi_{k-1}(t)-\phi_{k}(t))
\end{equation}
where  $T$ and $L_k$ denote the sampling time and the cell length respectively, whereas the flow
$\phi_k(t) = \min \left\{ d_k(\rho_k(t)), s_{k+1}(\rho_{k+1}(t)) \right\}$ between to adjacent locations $k$ and $k+1$ is computed as the minimum of the upstream \emph{demand} $d_k(\rho_k(t))$ of vehicles that seek to travel downstream and the downstream \emph{supply} $s_{k+1}(\rho_{k+1}(t))$ of free space. This model is called Cell Transmission Model (CTM) \cite{daganzo1994cell}. In the standard CTM, the flow is non-decreasing in the upstream density and non-increasing in the downstream density. However, there is empirical evidence of a capacity drop at a congested bottleneck, that is, the demand function $d_k(\cdot)$ slightly decreases as the upstream density exceeds the critical density. To deal with such capacity drop at critical density, we resort to a 2D representation of the fundamental diagram (see Fig. \ref{fig:fd_speedd_c}). We propose to estimate the capacity drop for bottleneck locations offline, using model-free Gaussian Process (GP) regressions \cite{rasmussen2004gaussian} to obtain a data driven estimate of the two-dimensional fundamental diagram.
\subsection{Low-level Control} 
\label{sec:low-level}
The main objective of the low-level control is to maximize local traffic flows by shifting the local traffic density towards the critical density, which is sufficient for close-to-optimal performance for a monotonic freeway. It can be achieved with the successful ALINEA algorithm \cite{papageorgiou1991alinea}, an integral feedback law in which the ideal metering rate is given as
\begin{align*}
\tilde r_k(t) := \hat r_k(t-1) + K_I \cdot (\rho^c_k - \rho_k(t)),
\end{align*}
where, $K_I$ is the integral gain chosen as in \cite{papageorgiou1991alinea} and $\hat r_k(t-1)$ is the on-ramp flow measured during the last sampling period. The only road parameter used by the feedback law is the critical density $\rho_k^c$, which is estimated online from data, as outlined before. However, the actual metering rate is subject to certain constraints. Obviously, it is non-negative and upper-bounded by some constant maximal on-ramp flow $\bar r_k$. We also allow for a user-defined (see Section \ref{Chap_Interface}) lower bound $\underline r_k$, which can be used to limit the maximal waiting time of drivers on the on-ramp. In addition, the space on the on-ramp is finite and it is paramount that the queue length $q_k(t)$ (in number of vehicles) does not exceed the maximal capacity $\bar q_k$ to avoid spill-back of the queue into adjacent arterial roads. Conversely, sometimes it may be required to hold back a certain amount of vehicles on the on-ramp to ease the traffic situation downstream, even if no congestion is imminent right at the on-ramp. To this end, we define a desired queue length $0 < q_k^*(t) \leq \bar q_k$, which will be chosen by the coordination algorithm as described in the following section. Thus, the actual metering rate $r_k(t)$ is saturated to the interval
\begin{align*}
\max \left\{ \underline r_k, \frac{1}{\Delta t} \left( q_k(t) - \bar q_k \right) + \hat d_k(t) \right\} \leq r_k(t) \leq \min \left\{ \bar r_k , \frac{1}{\Delta t} \left( q_k(t) - q_k^*(t) \right) + \hat d_k(t)  \right\} .
\end{align*}
Here, $\hat d_k(t)$ is the prediction of the traffic demand arriving at the on-ramp in the next time interval and $\Delta t$ is the sampling time. The states $\rho_k(t)$ and $q_k(t)$ are estimated from measurement streams using a standard Kalman filter \cite{Kalman1960}.

\subsection{Ramp Coordination} 
\label{sec:coordination}
The aim of coordinated ramp metering is to target inefficiencies that result from limited space on the on-ramps in conjunction with the non-monotonic behavior of a congested bottleneck. In the spirit of the proactive approach of the proposed platform, it is necessary to predict congestion. As described in Section \ref{Chap_Detection}, we predict congestion by learning patterns from historic, large data sets. Then, efficiency of the coordinated ramp metering scheme hinges mainly on the accuracy of the predictions made by CEP, while the coordination algorithm can be described as a simple finite state machine that reacts accordingly.
\begin{figure}[hbtp]
\centering
\subfloat[State diagram to determine the active local control algorithm.]{\label{fig:state_diagram_1}\includegraphics[width=0.35\textwidth]{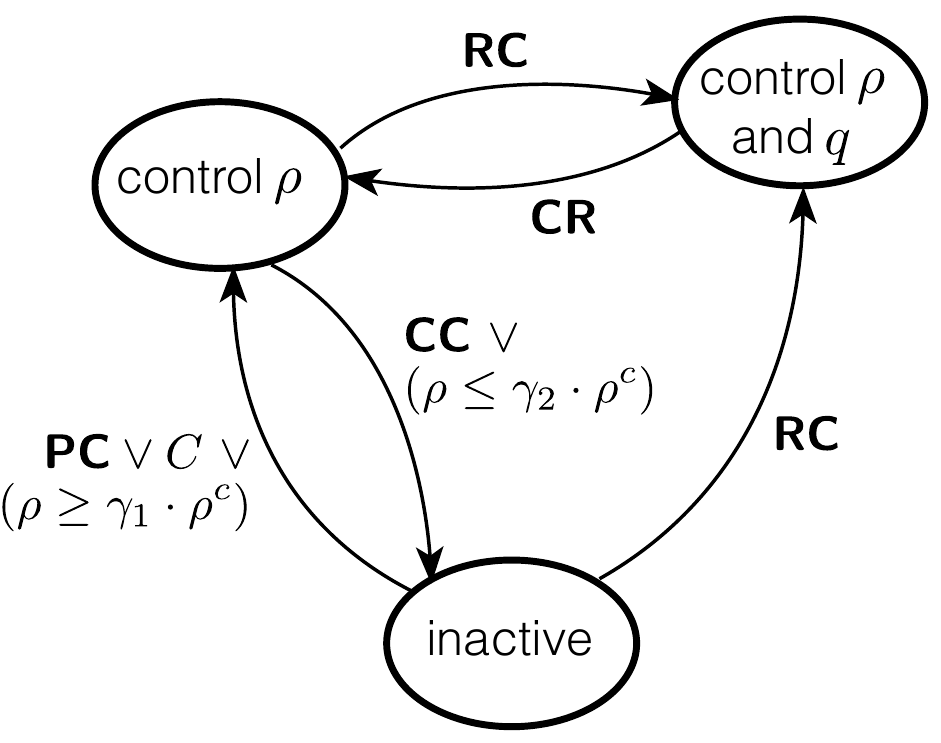}} \hspace{0.05cm}
\subfloat[Activation of upstream coordination.]{\label{fig:state_diagram_2}\includegraphics[width=0.25\textwidth]{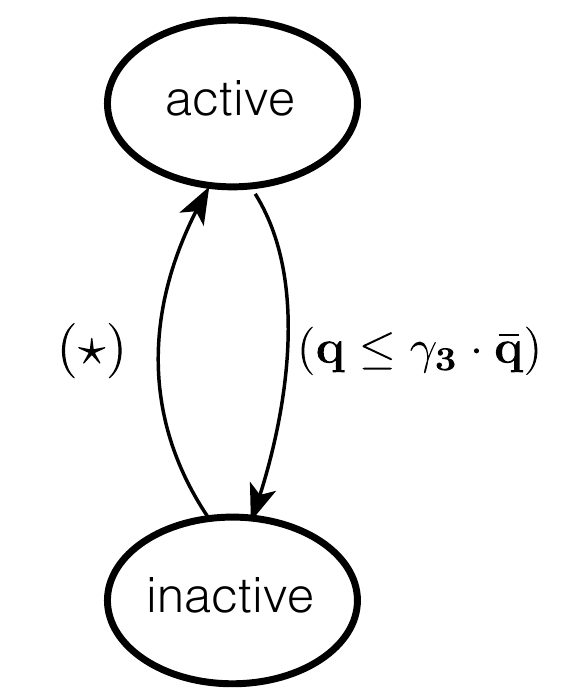}}
\caption{State diagrams of the coordination algorithm. The symbols and abbreviations are explained in the text.}
\label{fig:state_diagram}
\end{figure}
As depicted in Fig. \ref{fig:state_diagram_1}, the local feedback law described in Section \ref{sec:low-level} either controls the local density $\rho$ or both the local density and the on-ramp queue length $q$. In general, control of the density is activated if a congestion is detected and the queue length is controlled only if an upstream ramp requests coordination via the \emph{Ramp Coordination} (\textbf{RC}) event. If coordination is active between two ramps, we seek to balance the occupancies on both the upstream (us) and downstream (ds) on-ramp, that is, $q^*_{ds}(t) = \frac{\bar q_{ds}}{\bar q_{us}} \cdot q_{us}(t) $ \cite{papamichail2010heuristic}. Fig. \ref{fig:state_diagram_2} shows when a downstream ramp will request coordination from an upstream ramp. \textbf{RC} events are periodically sent if upstream coordination is active. Here, condition $(\star)$ is shorthand notation for $(\star) := {\bf ((TO_{PC} \vee TO_{PR}  \vee C) \wedge CA)} \vee (q \geq \gamma_4 \cdot \bar q) $, where the relevant events are \emph{Predicted Congestion} (\textbf{PC}), \emph{Predicted Ramp Overflow} (\textbf{PR}) and \emph{Congestion} (\textbf{C}). The Boolean variable \textbf{CA} is \textbf{true} if the local control algorithm is active, that is, it controls $\rho$, or $\rho$ and $q$ and the Boolean variables $\textbf{TO}_{PC/PR}$ are used to describe a trade-off with respect to the total expected travel time as described below. The remaining events \emph{Clear Congestion} (\textbf{CC}) and \emph{Clear Ramp Coordination}\footnote{This event is implemented as a \emph{Ramp Coordination} event with particular attribute values.} are used to determine when control can safely be deactivated. In both state diagrams in Fig. \ref{fig:state_diagram}, conditions in \textbf{bold} can be interpreted as the ``default" conditions. If these are comprised of events that might rarely not be predicted/ detected correctly, alternative conditions are also provided as a fall-back solution.
The parameters $\gamma_1 = 0.8, \gamma_2 = 0.7, \gamma_3 = 0.7$ and $\gamma_4 = 0.8$ are tuning parameters for defining thresholds for transitions.

Note that using multiple ramps comes at a cost. More cars are held back on the on-ramps, resulting in time lost for the drivers if it turns out that congestions would have been avoided without the use of coordination. Recall that the CEP engine also estimates the probability $\mathbb{P}[E]$ of a predicted event happening within some time horizon $T$. It is used to perform a trade-off between the (potential) benefits of preventing a congestion and the possibility of wasting driver's time on the on-ramps. To do so, we compare the additional waiting time $\Delta T_{ramp}$ on the upstream on-ramp to the time potentially wasted in congestion on the mainline $\Delta T_{ml}$ and define the Boolean trade-off variable $\text{TO}_{E} := \left( \mathbb{P}[E] \cdot \Delta T_{ml}' > (1- \mathbb{P}[E]) \cdot \Delta T_{ramp}' \right)$. Here, the event $E$ is either \textbf{PC} or \textbf{PR}, as both affect condition $(\star)$ in the same way. For a pair of ramps, the additional waiting time can be bounded by
\begin{align*}
\Delta T_{ramp} = \Delta t~ \cdot \sum_{t = 0}^{T / \Delta t} q_{us}(t) \leq T \cdot \frac{\bar q_{ds}}{\bar q_{ds} + \bar q_{us}} \bar q_{us} =: \Delta T_{ramp}' ,
\end{align*}
for a situation in which coordinated ramp metering was ultimately unnecessary and therefore, the total amount of cars stored on both ramps is less than the space available on the downstream ramp $q_{us}(t) + q_{ds}(t) \leq \bar q_{ds}$, for all $t \in [t,t+T]$. Conversely, if a congestion does arise within a time horizon $T$, the average, surplus demand is equal to at least $\Delta d \geq ( l \cdot (\rho^c_{ds} - \rho_{ds}(t) ) + \bar q_{ds} - q_{ds}(t) ) / T$. If  coordination is used, the surplus demand can also be stored on the upstream on-ramp, delaying the congestion by $\Delta T = \frac{\bar q_{us}}{\Delta d}$. Therefore, the additional time spent in congestion can be bounded by
\begin{align*}
\Delta T_{ml} \geq  \Delta T \cdot  \Delta \phi \cdot T_{con} = \frac{\bar q_{us} \cdot T }{ l \cdot (\rho^c_{ds} - \rho_{ds}(t) ) + \bar q_{ds} - q_{ds}(t) } \cdot \Delta \phi \cdot T_{con} := \Delta T_{ml}'.
\end{align*}
with $T_{con}$ the expected duration of the congestion and $ \Delta \phi$ the bottleneck capacity drop, which have to be estimated from historic data. Note that the inequality used in the trade-off provides a sufficient, but not a necessary condition for efficiency of coordination because only the bound $\Delta T_{ml}'$ and $\Delta T_{ramp}'$ can be computed. This is less restrictive than it might seem at first, since CEP will continue to produce updated predictions as traffic conditions evolve. Therefore, adopting a conservative approach at worst delays the usage of coordination slightly, until congestion can be predicted with sufficient accuracy.

\section{Dashboard design}\label{Chap_Interface}
While the automated event forecasting and decision making can address challenges relating to congestion and ramp metering, operators in the Road Traffic Control room, such as Grenoble's DIR CE are required to monitor and manage many other aspects of the road network.  Consequently, it was essential to integrate the output from the SPEEDD system into a visualisation which supported these other aspects of their work.  The goals of the Road Traffic Control room is to: maximize the available capacity of the road system, minimize the impact of incidents, manage demand regulation and congestion, assist in emergency service response, maintain public confidence in messages displayed on the VMS, and maintain a record of actions and events on the road system.
\par
\begin{figure}[hbtp]
\centerline{\includegraphics[scale = 0.55]{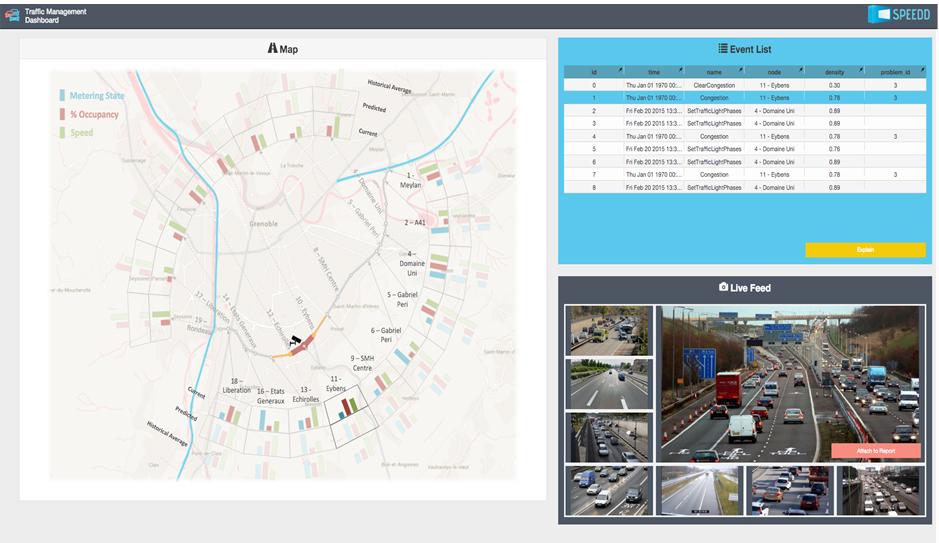}}
\caption{Developed traffic monitoring dashboard.}
\label{fig_UI}
\end{figure}

 Design of the dashboard for SPEEDD began with visits to DIR CE in which we spoke with and observed operators at work.  In addition to such observations, we were able to conduct eye-tracking studies in order to explore which information sources were most useful to the operators and what their information handling strategies were \cite{Starkeetal15,D8_1}.  Combining the material collected during field studies, we developed a set of Cognitive Work Analysis \cite{Vicente99} descriptions which provided different views on the operators' activity.  A dashboard which was implemented in the SPEEDD architecture and which resulted directly from our understanding of operators' decision making, information use and communications was produced as the first version of the dashboard. This was presented to operators and its usability explored. From this evaluation, a revised dashboard, depicted in Fig. \ref{fig_UI}, was developed. A detailed description of the design process and decisions made in the development process of the design can be found in \cite{D5_3}.

\par Fig. \ref{fig_UI} consists of three main areas.  In the top right of the screen, an event list provides a time-ordered set of outputs from the SPEEDD congestion prediction algorithm. This highlights to the operators event which need to be managed. Some of the events will be automatically handled by the ramp-metering and, as such, need not be brought to the operators' attention.  When ramp-metering is implemented, the status of the ramp is changed in the bar charts associated with each ramp on the map (on the left of the screen).  The map shows the Grenoble South Ring and, through colour-coding of the discrete segments of the road, indicating the current level of congestion.  The bar charts at each ramp also indicate the traffic flow through the ramp and the congestion adjacent to the ramp.  In the centre of the map, one can see a small camera icon.  This indicates the site of the CCTV camera which has produced the largest image in the CCTV panel (on the bottom right of the screen).  The CCTV panel shows a collection of images from the CCTVs which the operators can use to diagnose the level and possible causes of congestion.  The aim of the dashboard design was to produce a clear and simple overview of the road system, in order to allow operators to maintain a high level of situation awareness, and to allow them to see the decisions that the automated system was enacting (both in terms of the changes to ramps and display of congestion on the road, and in terms of those aspects of congestion which were not handled directly by ramp metering). For the events in the event list, the operator will indicate what action was taken, e.g., in terms of calling up VMS messages to alert drivers to congestion or to advise them to decrease their speed.  The operator actions can be combined with the initial changes in ramp metering, e.g., ramp metering initiated at a given time, in order to produce a detailed log of operations in the control room.

\section{Proactive Event-driven system architecture}\label{Chap_Architecture}

%

\begin{figure}[hbtp]
\centerline{\includegraphics[scale = 0.5]{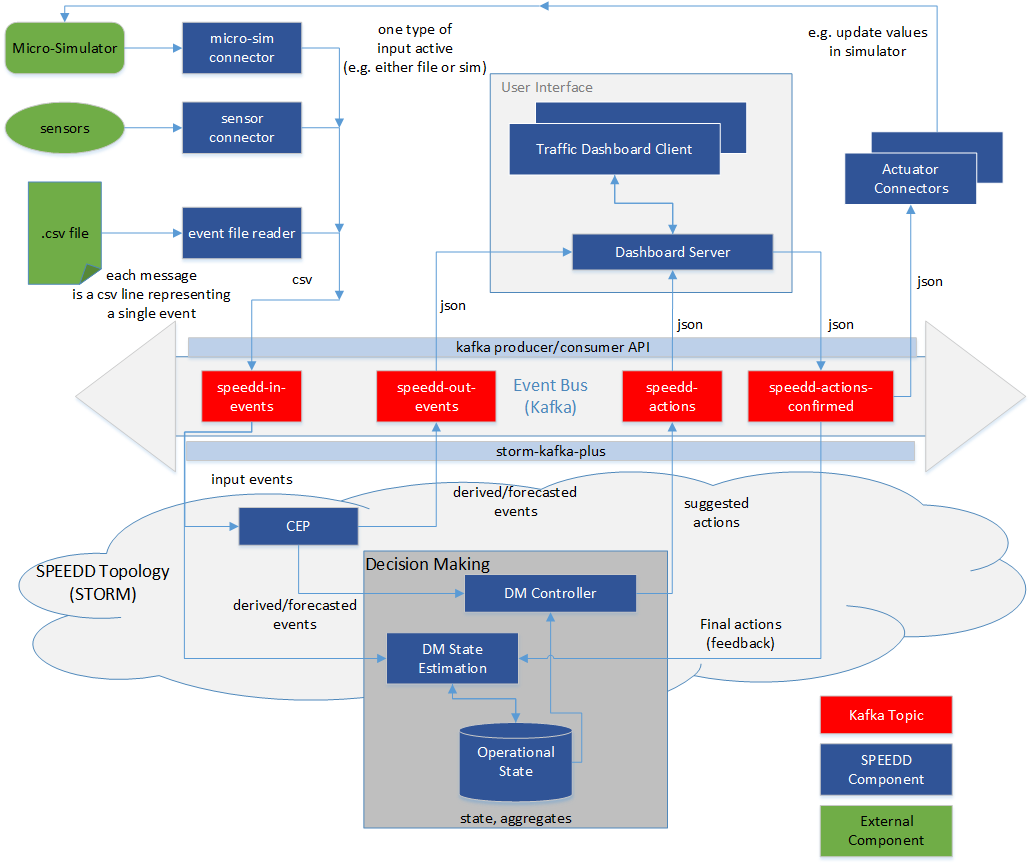}}
\caption{Run time architecture of the SPEEDD system.}
\label{fig_runtime_architecture}
\end{figure}

Fig. \ref{fig_runtime_architecture} shows the event-driven architecture run time represented as a group of loosely-coupled components interacting through events. The event bus serves as the communication and integration platform for the run time.
In general, input from the operational systems (traffic sensor readings) are represented as events and injected into the system by posting a new event message to the event bus. These events are consumed by complex event processing (CEP). The derived events representing detected or forecasted situations that the CEP component outputs are posted to the event bus as well. The decision making module listens to these events so that the decision making procedure is triggered upon a new event representing a situation that requires a decision. The output of the decision making represents the action to be taken to mitigate or resolve the situation. These actions are posted as action events. The visualization component (dashboard) consumes events coming from two sources: the situations (detected as well as forecasted) and the corresponding actions suggested by the automatic decision components. The user can accept the suggested action as it is, modify the suggested action's parameters, or reject it (and even decide upon a different action). In the case where an action is to be performed, the resulting action will be sent as a new event to the event bus so that the corresponding actuators are notified. The full description of the proposed proactive event-driven architecture can be found in \cite{Fournieretal2015}.

The proposed event-driven architecture can be run in an open, closed, or hybrid loop mode. In the current scenario, operators interact with the outputs of the prototype through a dashboard. The dashboard client communicates, via the dashboard server, with the modules of the architecture. Operators can accept, respond to, or make suggestions and control actions. Actions taken by operators via the dashboard are fed back into the run time as events, thus allowing for the seamless integration of expert knowledge and the outputs of complex algorithms.

It is worth noting that the Machine Learning for event definitions is not part of the run time architecture described in Fig. \ref{fig_runtime_architecture}. The automated construction of traffic congestion patterns (see Section \ref{Chap_Detection}) is performed at design time.
%
%

\section{Evaluation}\label{Chap_Evaluation}
In this section, we first evaluate our approach of learning traffic event definitions using Machine Learning techniques and then the different components of the proposed run time platform.
\subsection{Learning Event Definitions}
We applied \OSLa\ (see Section \ref{machine_learning}) to traffic management using real data from the magnetic sensors mounted on the Grenoble South Ring, consisting of approximately $3.3$GiB of sensor readings (one month data). Annotations of traffic congestion are provided by human traffic controllers, but only very sparsely. To deal with this issue, we also used a synthetic dataset generated by the traffic micro-simulator of GTL (see Section \ref{Chap_scenario}). The synthetic dataset concerns the same location and consists of 6 simulations of one hour each ($\approx 18.6$MiB). 

A set of first-order logic functions is used to discretize the numerical data (speed, occupancy) and produce input events such as, for instance, $\pHappensAt{\mathtt{fast{\char`_}Slt55}(\mathtt{53708})}{\mathtt{100}}$, representing that the speed in the fast lane of location $53708$ is less than 55 km/hour at time $100$. The total length of the training sequence in the real data case consists of $172,799$ time-points, while in the synthetic data it consists of 238 time-points.
The evaluation results were obtained using MAP inference \cite{huynh2009max} and are presented in terms of $F_1$ score. In the real dataset, all reported statistics are micro-averaged over the instances of recognized CEs using 10-fold cross validation over the entire dataset, using varying batch sizes. At each fold, an interval of $17,280$ time-points was left out and used for testing. In  the synthetic data, the reported statistics are micro-averaged using 6-fold cross validation over 6 simulations by leaving one out for testing. 

\begin{figure}[htbp]
\centerline{
\includegraphics[width=0.43\textwidth]{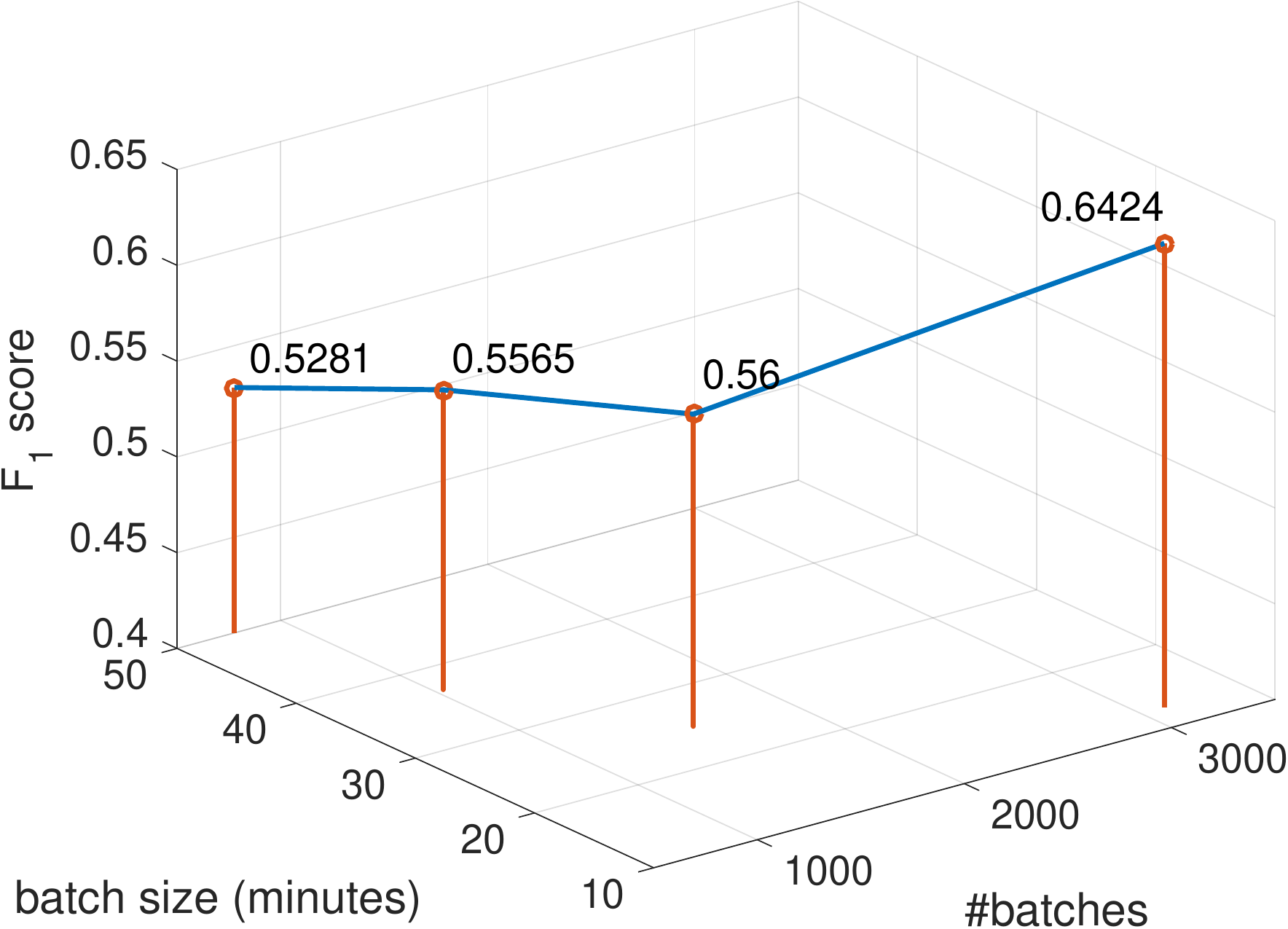}~~
\includegraphics[width=0.43\textwidth]{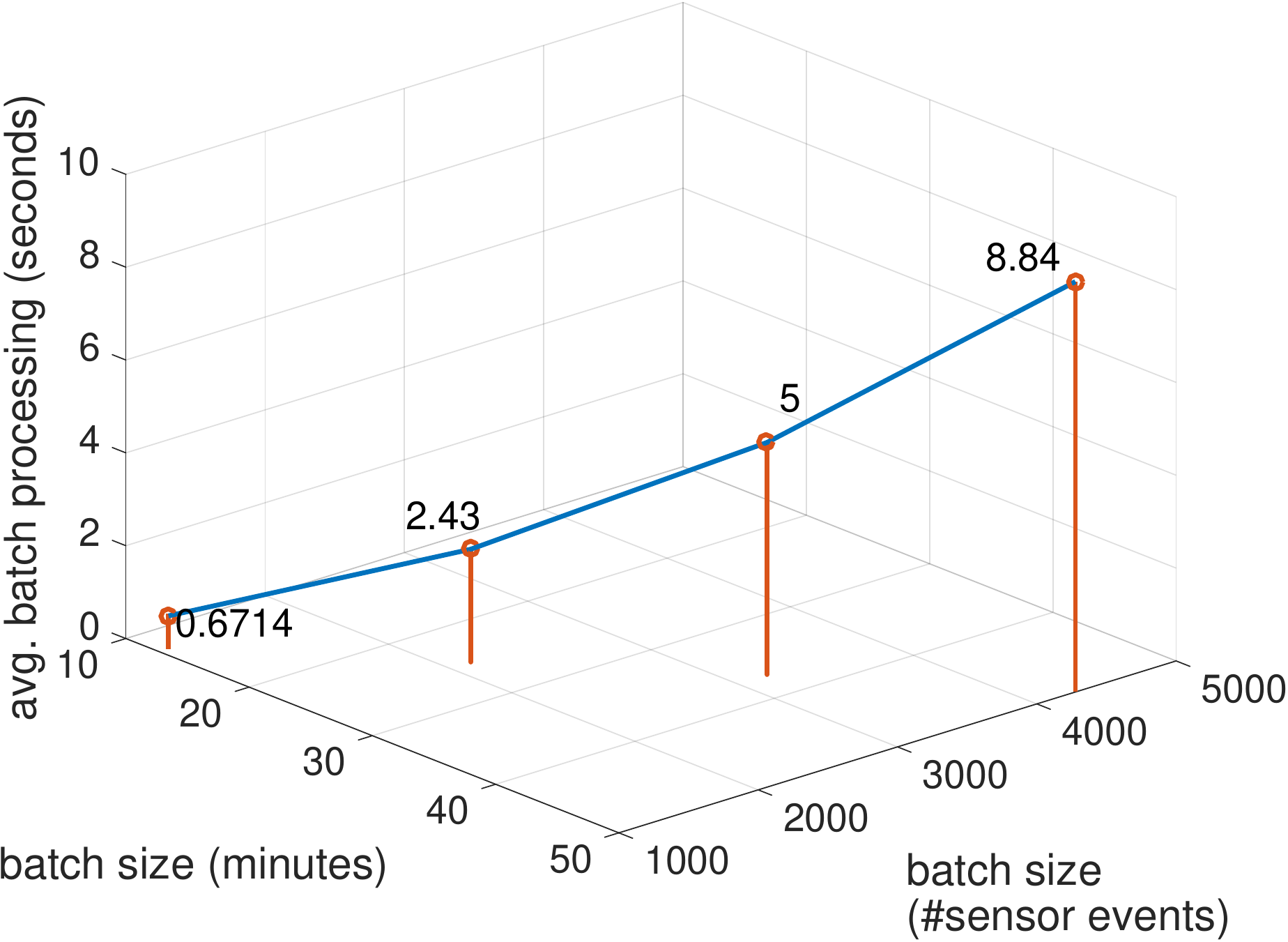}
}\vspace{0.2cm}
\centerline{
\includegraphics[width=0.43\textwidth]{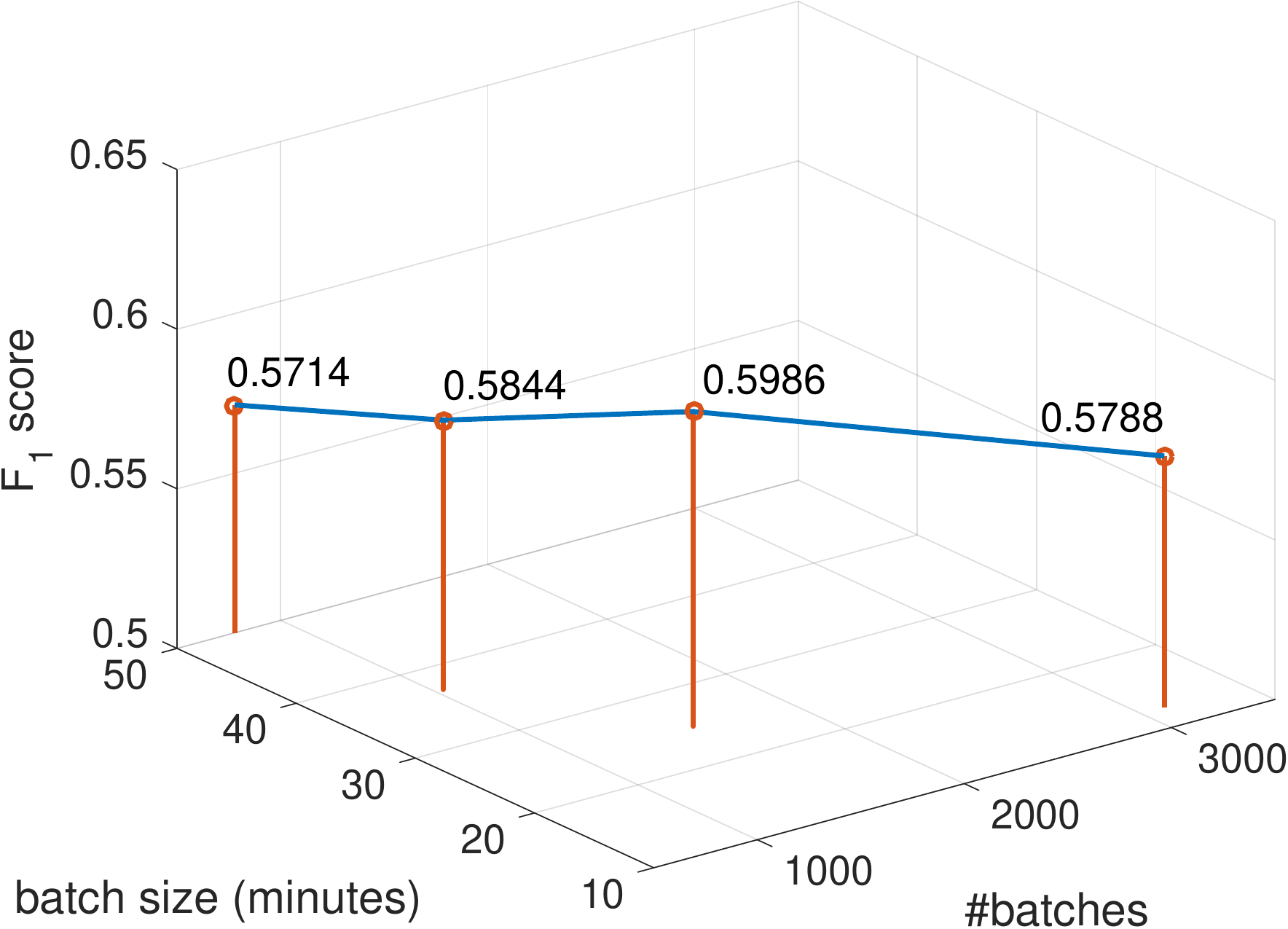}~~
\includegraphics[width=0.43\textwidth]{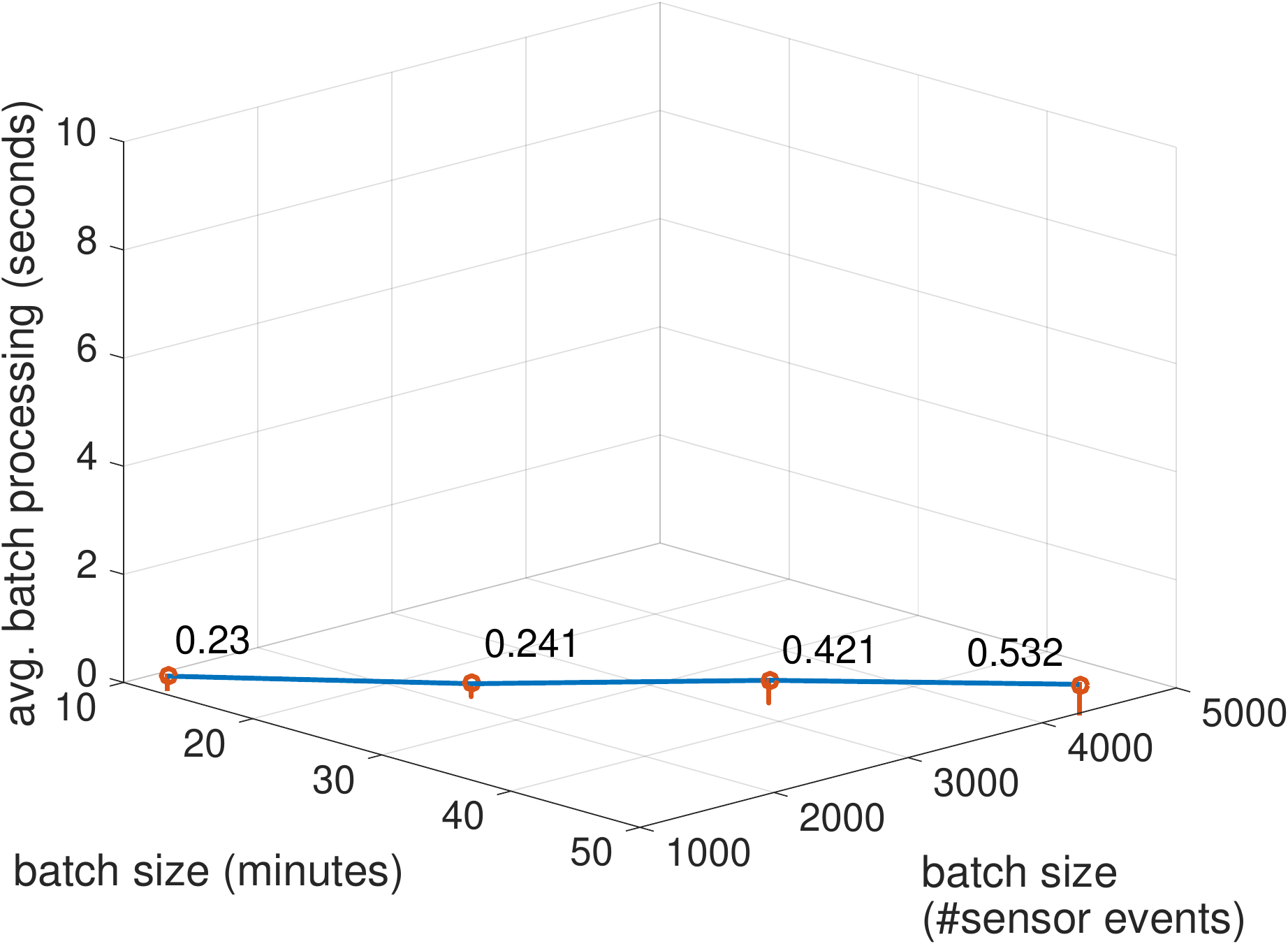}
}
\caption{Real dataset: $F_1$ score (left) and average batch processing time (right) for \OSLa\  (top), and AdaGrad operating on manually constructed traffic congestion rules (bottom). In the left figures, the number of batches (see the $Y$ axes) refers to number of learning steps.}
\label{fig:nokb_results_real}
\end{figure}

Fig. \ref{fig:nokb_results_real} presents the experimental results on the real dataset. We compare \OSLa\ against the AdaGrad online weight learner \cite{duchi2011AdaGrad} that optimizes the weights of a manually constructed traffic congestion definition.
The predictive accuracy of the learned models, both for \OSLa\ and AdaGrad, is low. This arises mainly from the largely incomplete supervision.
In \OSLa, the predictive accuracy increases (almost) monotonically as the learning steps increase.
On the contrary, the accuracy of AdaGrad is more or less constant.
\OSLa\ outperforms AdaGrad in terms of accuracy. (\OSLa\ achieves a $0.64$ $F_1$ score, while the best score of AdaGrad is $0.59$.) This is a notable result.
The absence of proper supervision penalizes the hand-crafted rules, compromising the accuracy of AdaGrad that uses them. \OSLa\ is not penalized in this way, and is able to construct rules with a better fit in the data, given enough learning steps. For some locations of the highway, \OSLa\ has constructed rules with different thresholds for speed and occupancy than those of the hand-crafted rules. With respect to efficiency (see the right diagrams of Fig. \ref{fig:nokb_results_real}), unsurprisingly AdaGrad is faster and scales better to the increase in the batch size. At the same time, \OSLa\ processes data batches efficiently, much faster than their duration. For example, \OSLa\ takes less than 9 sec to process a 50-minute batch including $4,220$ sensor readings.

\begin{figure}[htbp]
\centerline{
\includegraphics[width=0.43\textwidth]{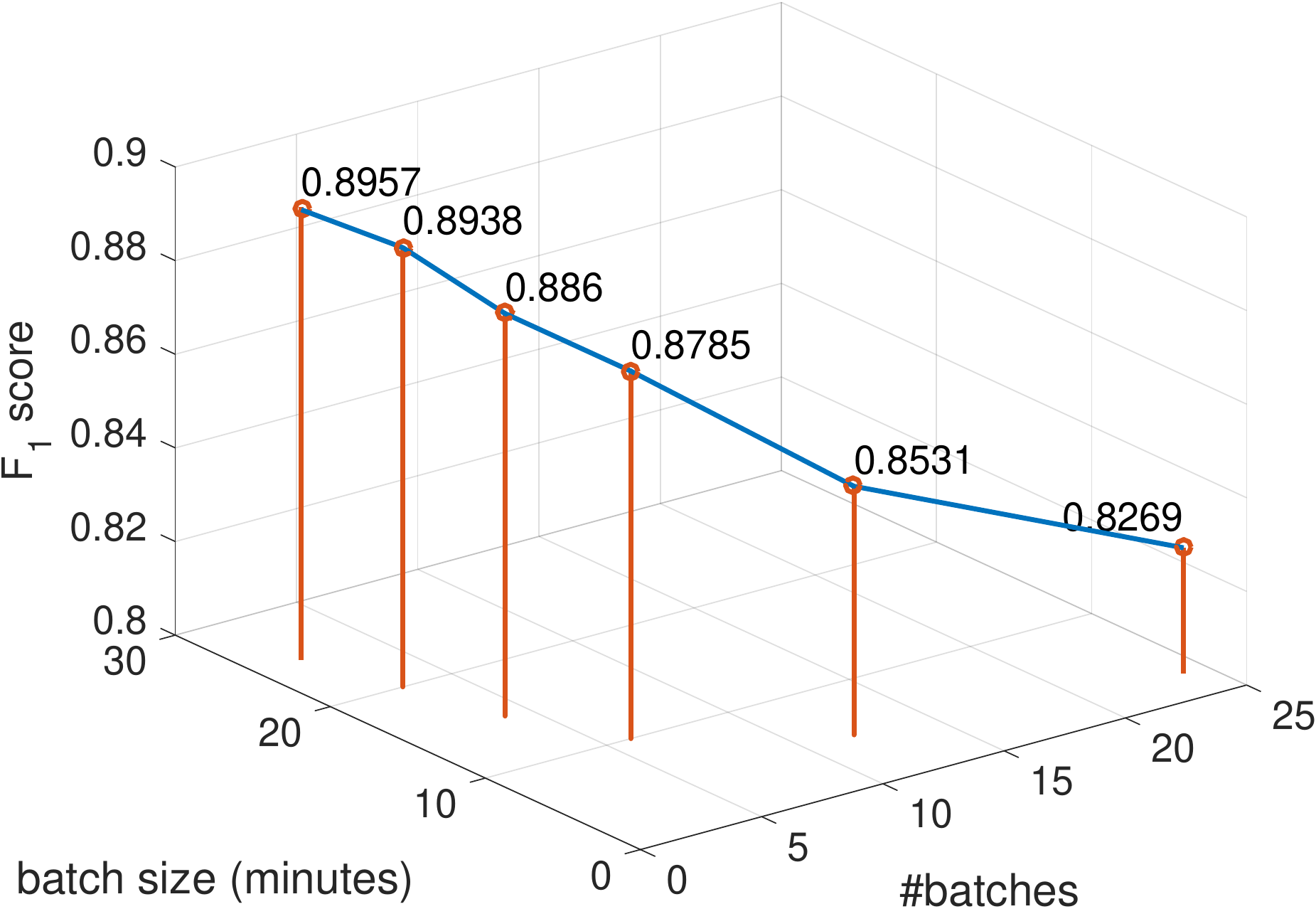}~~
\includegraphics[width=0.43\textwidth]{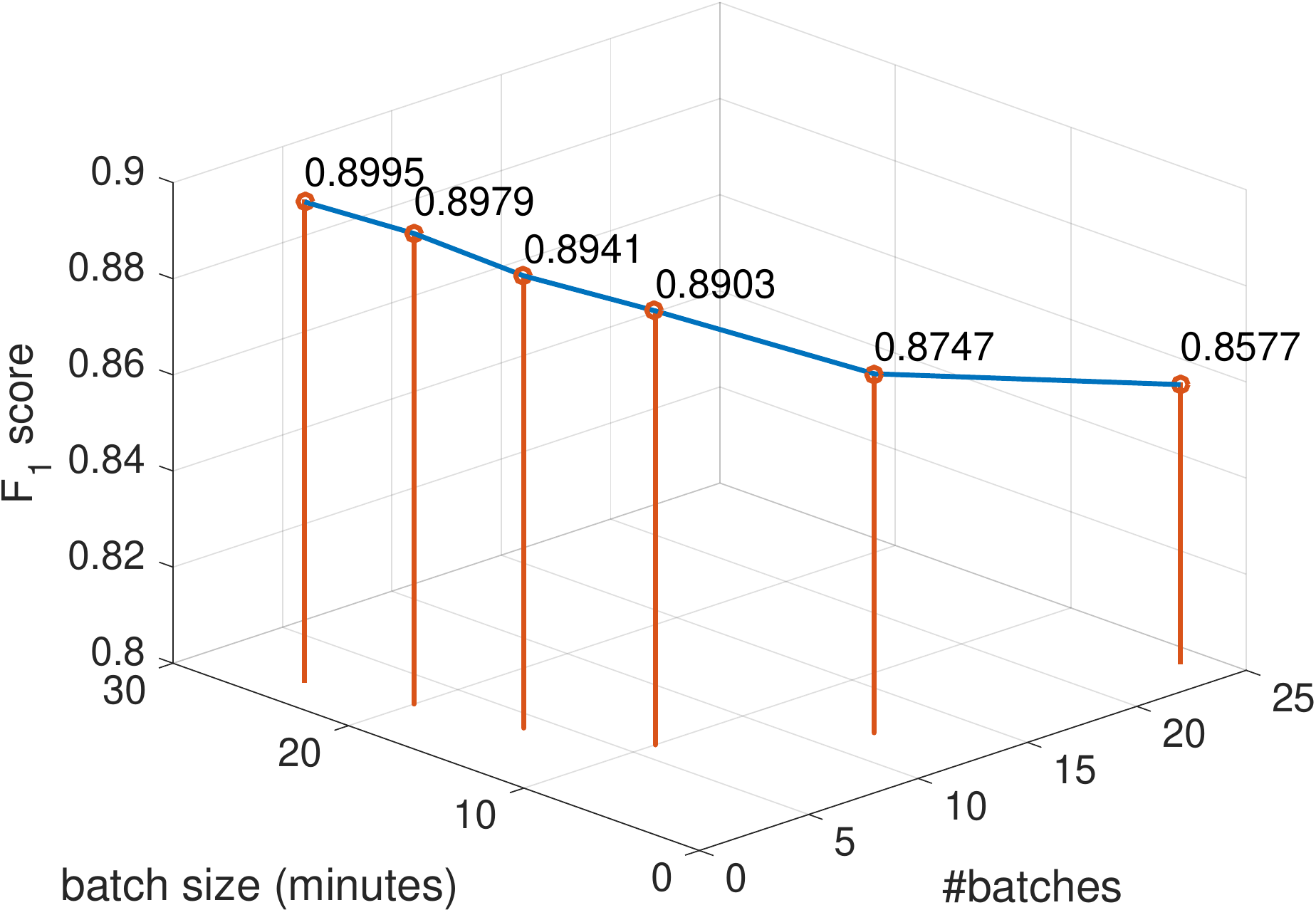}
}
\caption{Synthetic dataset: $F_1$ score for \OSLa\ (left) and AdaGrad operating on manually constructed traffic congestion rules (right).}
\label{fig:nokb_results_synthetic_speed_only}
\end{figure}

To test the behavior of \OSLa\ under better supervision, we made use of a synthetic dataset produced by the traffic micro-simulator of GTL. Fig. \ref{fig:nokb_results_synthetic_speed_only} presents the experimental results. Not surprisingly, the predictive accuracy of the learned models in these experiments is much higher as compared to real dataset. Moreover, the accuracy of \OSLa\ and AdaGrad is affected mostly by the batch size: accuracy increases as the batch size increases. The synthetic dataset is smaller than the real dataset and thus, as the batch size decreases, the number of learning steps is not large enough to improve accuracy.
The best performance of \OSLa\ and AdaGrad is almost the same (approximately $0.89$). In other words, \OSLa\ can match the performance of techniques taking advantage of  rules crafted by human experts. This is another notable result.

\subsection{Event forecasting} In order to explore the quality of our CEP module, we ran a test comprising of 20 simulations generated by the traffic micro-simualtor of GTL along with annotations of congestions. The annotations of congestions include the location and the time the congestion is detected.
First, we evaluated the quality of our Congestion pattern against the annotated data. We checked the proportion of detections by our EPA that were annotated in the data as congestions (precision) and second, the proportion of congestions we were able to detect out of all the annotated congestions (recall). In all our simulations our precision was $100\%$, while the average recall over all the simulations was $72\%$. This can be easily explained: the rule implemented has been given to us by the domain expert, who is the one to identify the congestions in the simulations, thus giving a perfect precision. However, when implementing the pattern we applied a ``stricter" criterion for the rule than the one in the simulator: we took into account not just the average speed critical thresholds, but also density thresholds, therefore we have a less success rate in the recall of the results, i.e., there were annotations of congestion in the data that we ``missed". 

As a second step, we aimed at checking a more interesting question, that is, whether the inclusion of uncertainty aspects enables us to predict a congestion in the highway before it reaches critical thresholds, as opposed to detecting it once it happens. We addressed this question by having two EPNs, once including uncertainty aspects and the other one without uncertainty, i.e. deterministic; and running the tests twice, one time for each EPN (with and without uncertainty). This is a common approach in CEP engines dealing with uncertainty (see for example in \cite{cugola}). The deterministic case served as baseline, as we knew at this stage that all our congestions have been detected correctly. The precision of our results indicates the proportion of congestions we were able to predict (in other words, PredictedCongestion pointed out correctly to a congestion), whereas the recall indicates the proportion of congestions we were able to detect out of all the annotated congestions (in other words, PredictedCongestion pointed out correctly out of all congestions). We used a threshold of 0.6 in the certainty attribute to determine whether to consider PredictedCongestion as a congestion. In other words, only PredictedCongestion alerts with a certainty value larger than 0.6 were considered in our calculations of precision and recall. In these tests, the average precision was $91\%$ and the average recall was $75\%$. Furthermore, PredictedCongestion event is emitted $3$ to $4$ minutes before a Congestion is detected, thus enabling the system to take proactive actions in order to alleviate these congestions. The recall average indicates that there are other situations that cause congestions which are not detected by our pattern. Further analysis shows that these situations are characterized by ``jumping data", meaning, the values of speed and density tend to jump thus not satisfying the increasing build-up which is required in our pattern. We are currently investigating these ``jumping" cases to see if we can identify some common behavior/pattern.

\subsection{Decision making}
\begin{figure}[hbtp]
\centering
\subfloat[Density for local control.]{\label{fig:dens_cl}\includegraphics[width=0.48\textwidth]{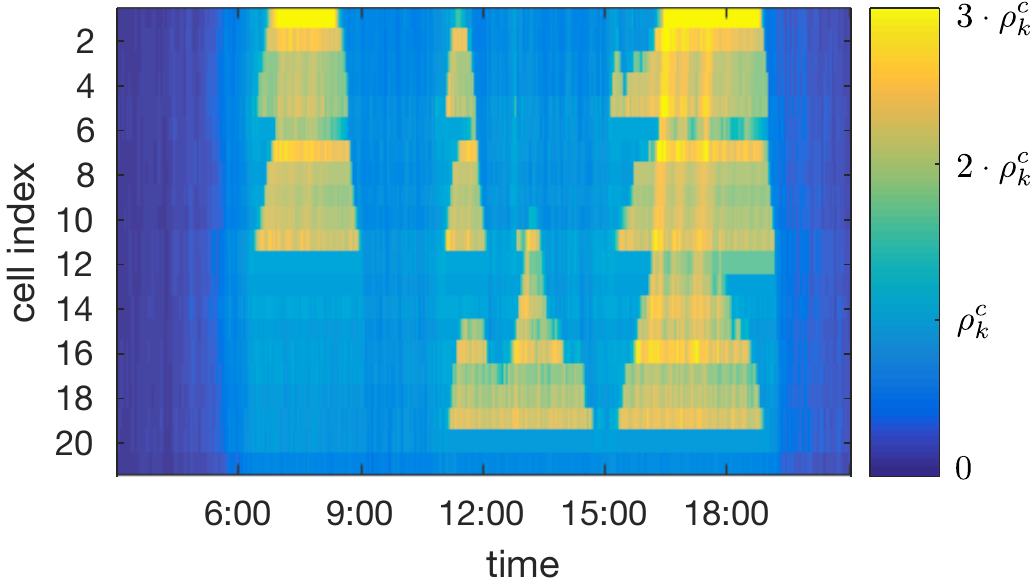}} \quad
\subfloat[Queue occupancies for local control.]{\label{fig:ramp_cl}\includegraphics[width=0.48\textwidth]{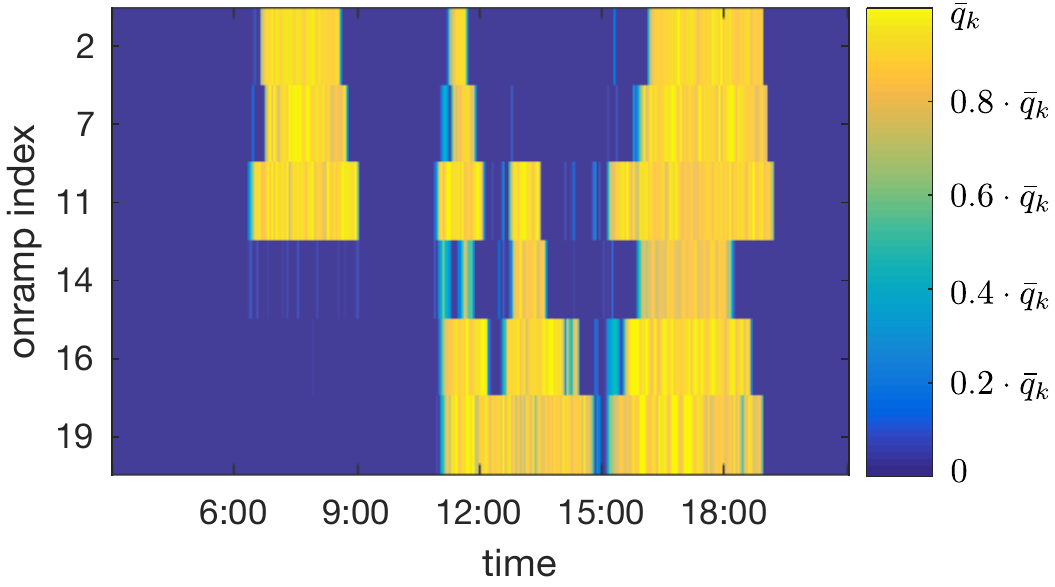}} \\
\subfloat[Density for coordinated control.]{\label{fig:dens_co}\includegraphics[width=0.48\textwidth]{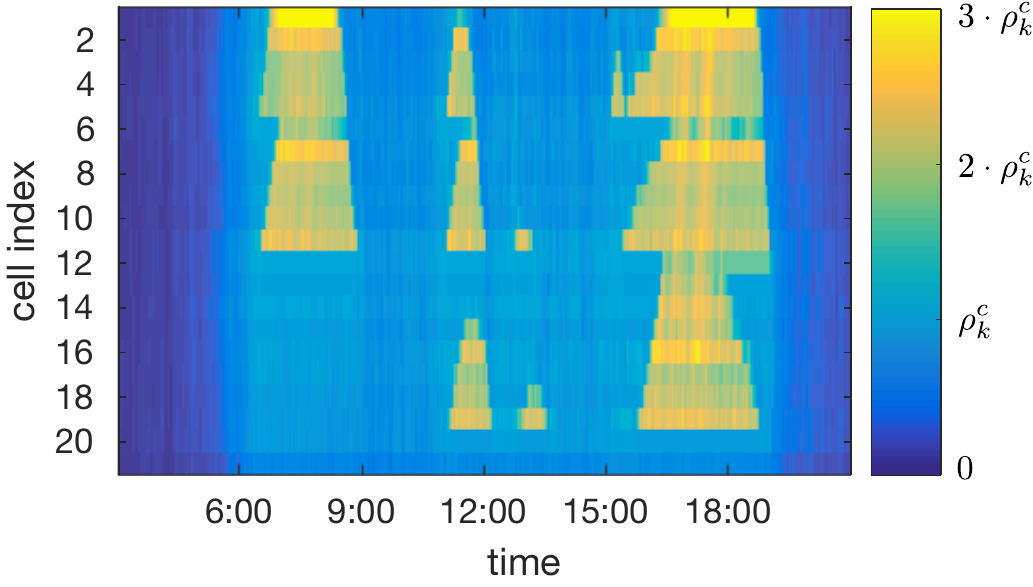}} \quad
\subfloat[Queue occupancies for coordinated control.]{\label{fig:ramp_co}\includegraphics[width=0.48\textwidth]{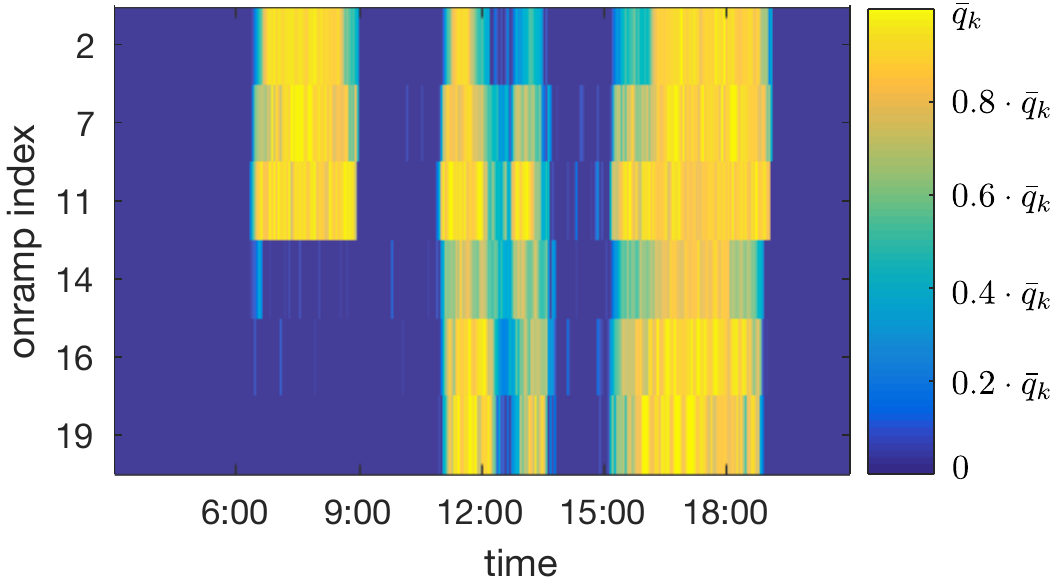}}
\caption{Simulation results for traffic demands of April 19$^{\text{th}}$, 2014. Coordination distributes vehicles among on-ramps, thereby reducing traffic on the mainline and increasing the bottleneck flows, in particular for cell $11$ and $19$.}
\label{fig:simulation}
\end{figure}

We evaluated the decision making module by considering that the ramps with indices $k \in \{2, 6, 7, 8, 9 \}$ as depicted in Figure \ref{fig_GTL} are used for ramp metering. We assume that on-ramp queues are extended to provide storage space for up to $50$ cars each. The simulation is conducted as described in \cite{schmitt2016sufficient} with non-monotonic demand functions. To quantify the benefits of ramp metering, we use the Total Time Spent (TTS), a standard metric defined as the sum of the travel times of all cars for a certain day. We perform three types of simulations. First, we simulate traffic without ramp metering to obtain a baseline performance, $\text{TTS}_{ol}$. Second, simulations using local ramp metering as described in Section \ref{sec:low-level} are performed, but no coordination between ramps is used. The corresponding travel time is denoted $\text{TTS}_{cl}$. Third, we employ coordinated ramp metering with the coordination along the lines of Section \ref{sec:coordination} and denote the corresponding total time spent as $\text{TTS}_{co}$. The parameters of the coordination are chosen as $\gamma_3 = 0.1$ and $\gamma_4 = 0.2$. For the five-week period, we obtain relative savings of
\begin{equation*}
\frac{\text{TTS}_{ol} - \text{TTS}_{cl}}{\text{TTS}_{ol} } = 9.9\% \quad\text{and} \quad \frac{\text{TTS}_{ol} - \text{TTS}_{co}}{\text{TTS}_{ol} } = 13.6\% .
\end{equation*}
Benefits of coordination tend to increase as traffic demand increases, while conversely, no benefits are obtained on days with no or only light congestion for an uncontrolled freeway. However, TTS does not only quantify time wasted in congestion and in on-ramp queues, but vehicles traveling at free-flow velocity contribute significantly as well. Ramp metering cannot provide any benefits during times at which the uncontrolled freeway is not congested. Therefore, we define the Total-Free-flow-Time TFT as the travel time accumulated by all vehicles on a hypothetical freeway, that is always uncongested, i.e.\, all vehicles travel at free-flow velocity at all times. The relative savings in terms of time wasted in congestion and in on-ramp queues for all days amount to
\begin{equation*}
\frac{\text{TTS}_{ol} - \text{TTS}_{cl}}{\text{TTS}_{ol}- \text{TFT}} = 25.2\% ~\text{and} \quad \frac{\text{TTS}_{ol} - \text{TTS}_{co}}{\text{TTS}_{ol} - \text{TFT}} = 34.6\% .
\end{equation*}
The results are visualized in Figure \ref{fig:simulation} for one day that provides average savings\footnote{April 19$^{\text{th}}$, 2014: $\frac{\text{TTS}_{ol} - \text{TTS}_{cl}}{\text{TTS}_{ol} \quad \text{TFT}} = 21.2\%$, $\frac{\text{TTS}_{ol} - \text{TTS}_{co}}{\text{TTS}_{ol} - \text{TFT}} = 33.8\%$ }. Note that these savings are larger than the ones reported in \cite{schmitt2016sufficient} and \cite{GomesHorowitz2006}. This is no surprise, since these papers use the monotonic CTM. In a monotonic setting, the only benefit of ramp metering is to avoid blocking off-ramps with spilled-back mainline congestion. It shall be noted however that both the time spent in congestion and the relative savings of coordination seem to be sensitive to the traffic demands. In a non-monotonic setting, small changes in demands may cause large differences in open- or closed-loop behavior. As stated earlier, coordination tends to provide larger relative savings for more severe congestion.

\subsection{Evaluation of dashboard design}
\begin{figure}[hbtp]
\centerline{\includegraphics[scale = 0.65]{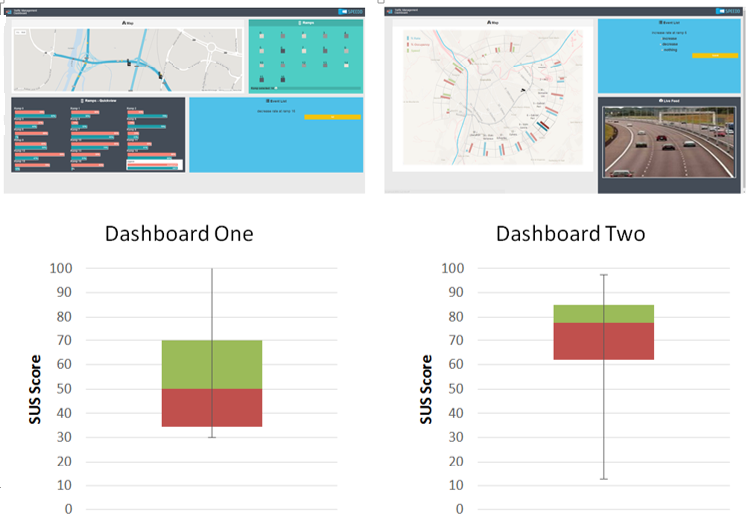}}
\caption{Comparison of the original dashboard (one) with the revised dashboard (two) in terms of subjective rating of usability.}
\label{fig_dashboard_evaluation}
\end{figure}

As part of the usability evaluation of the dashboard design, participants ($N=24$) completed a series of tasks using the original dashboard and the revised version which was described in Section \ref{Chap_Interface}. Participants were asked to complete the Software Usability Scale (SUS) questionnaire \cite{Brooke96} for each dashboard. The SUS questionnaire consists of 10 simple questions concerning the potential usefulness and benefit that users feel that the dashboard might provide them with. Each statement is rated on a scale of 0 to 4. The scoring of responses then involves subtracting 1 from odd-numbered questions and subtracting scores of even-numbered questions from 5.  This is because the questions alternate between positive and negative connotations.  Scores are then summed and multiplied by 2.5, to give a final score out of 100.  As a rule of thumb, scores in excess of 65 are deemed ``acceptable".  Fig. \ref{fig_dashboard_evaluation} compares the evaluation of the dashboard described in section 5 with the original version: the median scores were 49 for the original version (indicating that the design was of lower than acceptable usability) and 69 for the revised version (indicating that participants felt the design to be acceptable).

In addition to collecting subjective opinion of the usability of the two dashboard designs, an experiment was conducted in which 24 participants each completed $60$ ramp metering and traffic congestion tasks (e.g. task involving decisions on whether to alter the ramp metering rate and whether level of congestion has changed on different sections of the road). The tasks were completed with each dashboard and under different levels of automation reliability (low = $20\%$; medium = $50\%$ and high = $80\%$).  This latter condition was introduced to explore how users might respond to recommendations which were based on noisy or incomplete data (hence resulting in erroneous advice). Analysis of decision time showed that responses were significantly faster with dashboard 2 (mean decision time approximately $14.25$s for dashboard 1 and $10.5$s for dashboard 2) and also varied with reliability (mean decision time approximately $13.0$s for low; $12.75$s for medium; $11.5$s for high). In terms of decisions, users were able to match the reliability of the automation, i.e. when the reliability levels were low and medium, users would only respond to the ``correct" answers and were able to compensate for the errors to some extent.  However, decision accuracy for the low and medium reliability automation was $\leq 85\%$, and for the high reliability was around $92\%$.  Thus, when the automation performed poorly, human decision making could be affected.

\section{Conclusions}\label{Chap_Conclusions}
We have presented an intelligent platform for traffic management which includes a new ramp metering coordination scheme in the decision making module, an efficient dashboard for interacting with human operators, machine learning tools for learning event definitions and complex event processing tools able to deal with uncertainties inherent to the traffic use case. It has been shown that the developed machine learning tool can match the performance of techniques taking advantage of  rules crafted by human experts while complex event processing tools are able to predict congestion $3$ to $4$ minutes before Congestion happens even with uncertain and noisy data. The decision making module using coordinated ramp metering improves Total Spent Time compared to ramp metering without coordination using the current standard local feedback algorithm ALINEA. It is worth noting that even though the system is able to take proactive actions in order to alleviate congestions, the recall average indicates that there are other situations that cause congestions which are not detected by our patterns. Hence, there remains a need to ensure an integrated human-automation decision system as we have implemented in SPEEDD. Future works also include analysis and definition of patterns exhibiting jumps rather than trends.

\section*{Acknowledgments}
This  work  is part  of the  EU-funded  SPEEDD  project  (FP7-ICT 619435). We would also thank Philippe Mansuy of DIR CE for giving us access to the Grenoble South Ring Control Room and valuable comments of traffic operators.

\bibliographystyle{plain}
\bibliography{acmsmall-sample-bibfile,references,bibliography_dm,bibliography_dm_eval}

\begin{thebibliography}{10}

\bibitem{Artikis2012}
A.~Artikis, O.~Etzion, Z.~Feldman, and F.~Fournier.
\newblock {Event Processing under Uncertainty}.
\newblock In {\em {DEBS 2012}}, 2012.

\bibitem{D5_3}
C.~Baber, S.~Starke, X.~Chen, N.~Morar, and A.~Howes.
\newblock The design of user interfaces for the speedd prototype,3rd report.
\newblock Technical Report FP7-619435/SPEEDD-D5.3, EU, Scalable Data Analytics,
  Scalable Algorithms, Software Frameworks and Visualization ICT-2013 4.2.a,
  2016.

\bibitem{BovySalomon2002}
P.~Bovy and I.~Salomon.
\newblock {Congestion in Europe: Measurements, patterns and policies.}
\newblock {\em {Travel behaviour: Spatial patterns, congestion and modelling}},
  pages 143--179, 2002.

\bibitem{Brooke96}
J.~Brooke.
\newblock {SUS: a quick and dirty usability scale.}
\newblock In P.W. Jordan, B.~Weerdmeester, B.A. Thomas, and I.L. McLelland,
  editors, {\em Usability Evaluation in Industry}, pages 189--194. Taylor and
  Francis, London, 1996.

\bibitem{D8_1}
C.~Canudas~de Wit, I.~Bellicot, F.~Garin, P.~Grandinetti, A.Y. Kibangou,
  F.~Morbidi, M.~Schmitt, A.~Hempel, C.~Baber, and N~Cooke.
\newblock User requirements and scenario definition.
\newblock Technical Report FP7-619435/SPEEDD-D8.1, EU, Scalable Data Analytics,
  Scalable Algorithms, Software Frameworks and Visualization ICT-2013 4.2.a,
  2014.

\bibitem{canudasdewit:hal-01059126}
C.~Canudas~de Wit, F.~Morbidi, L.~Leon~Ojeda, A.Y. Kibangou, I.~Bellicot, and
  P.~Bellemain.
\newblock {Grenoble Traffic Lab: An experimental platform for advanced traffic
  monitoring and forecasting}.
\newblock {\em {IEEE Control Systems}}, 35(3):23--39, 2015.

\bibitem{cugola}
G.~Cugola, A.~Margara, M.~Matteucci, and G.~Tamburrelli.
\newblock {Introducing uncertainty in complex event processing: model,
  implementation, and validation}.
\newblock {\em {Computing}}, pages 1--42, 2014.

\bibitem{daganzo1994cell}
C.F. Daganzo.
\newblock The cell transmission model: A dynamic representation of highway
  traffic consistent with the hydrodynamic theory.
\newblock {\em Transportation Research Part B: Methodological}, 28(4):269--287,
  1994.

\bibitem{Downs2004}
A.~Downs.
\newblock {Why Traffic Congestion is Here To Stay $...$ and Will Get Worse}.
\newblock {\em {ACCESS Magazine}}, 25(1), 2004.

\bibitem{duchi2011AdaGrad}
J.~Duchi, E.~Hazan, and Y.~Singer.
\newblock {Adaptive Subgradient Methods for Online Learning and Stochastic
  Optimization}.
\newblock {\em Journal of Machine Learning Research}, 12:2121--2159, July 2011.

\bibitem{Engel2012}
Y.~Engel, O.~Etzion, and Z.~Feldman.
\newblock {A Basic Model for Proactive Event-Driven Computing}.
\newblock In {\em {DEBS 2012}}, pages 107--118, 2012.

\bibitem{Etzion2010}
O.~Etzion and P.~Niblett.
\newblock {\em {Event Processing in Action}}.
\newblock Manning Publication, 2010.

\bibitem{Fournieretal2015}
F.~Fournier, A.~Kofman, I.~Skarbovsky, and Skarlatidis A.
\newblock {Extending Event-Driven Architecture for Proactive Systems}.
\newblock In {\em {Proc. of Event Processing, Forecasting and Decision-Making
  in the Big Data Era (EPForDM), EDBT/ICDT Workshops.}}, pages 104--110, 2015.

\bibitem{GomesHorowitz2006}
G.~Gomes and R.~Horowitz.
\newblock {Optimal freeway ramp metering using the asymmetric cell transmission
  model}.
\newblock {\em {Transportation Research Part C: Emerging Technologies}},
  14(4):244--262, 2006.

\bibitem{huynh2009max}
T.N. Huynh and R.J. Mooney.
\newblock {Max-Margin Weight Learning for Markov Logic Networks}.
\newblock In {\em Proceedings of the European Conference on Machine Learning
  and Principles and Practice of Knowledge Discovery in Databases (ECML PKDD)},
  volume 5781 of {\em Lecture Notes in Computer Science}, pages 564--579.
  Springer, 2009.

\bibitem{huynh2011osl}
T.N. Huynh and R.J. Mooney.
\newblock {Online Structure Learning for Markov Logic Networks}.
\newblock In {\em Proceedings of the European Conference on Machine Learning
  and Principles and Practice of Knowledge Discovery in Databases (ECML-PKDD
  2011)}, volume~2, pages 81--96, September 2011.

\bibitem{Kalman1960}
R.E. Kalman.
\newblock A new approach to linear filtering and prediction problems.
\newblock {\em Journal of Basic Engineering}, 82(35), 1960.

\bibitem{Kojima1999}
M.~Kojima, C.~Nowakowski, and P.~Green.
\newblock Organization and structure of traffic management centers: Two case
  studies in {Michigan}.
\newblock Technical report, University of Michigan, Ann Arbor, MI, USA, 1999.

\bibitem{kowalksi1986EC}
R.~Kowalski and M.~Sergot.
\newblock {A Logic-based Calculus of Events}.
\newblock {\em {New Generation Computing}}, 4(1):67--95, 1986.

\bibitem{lighthill1955kinematic}
M.J. Lighthill and G.B. Whitham.
\newblock On kinematic waves ii. a theory of traffic flow on long crowded
  roads.
\newblock {\em Proceedings of the Royal Society of London A: Mathematical,
  Physical and Engineering Sciences}, 229(1178):317--345, 1955.

\bibitem{Luckham:2012}
D.C. Luckham.
\newblock {\em Event Processing for Business: Organizing the Real-Time
  Enterprise}.
\newblock John Wiley and sons, Inc, New jersey, USA, 1st edition, 2012.

\bibitem{vagmcs2016osla}
E.~Michelioudakis, A.~Skarlatidis, G.~Paliouras, and A.~Artikis.
\newblock {Online Structure Learning using Background Knowledge
  Axiomatization}.
\newblock In {\em Proceedings of the European Conference on Machine Learning
  and Principles and Practice of Knowledge Discovery in Databases (ECML-PKDD
  2016)}, volume~1, pages 242--237, September 2016.

\bibitem{mueller2008}
E.T. Mueller.
\newblock {Event Calculus}.
\newblock In {\em Handbook of Knowledge Representation}, volume~3 of {\em
  Foundations of Artificial Intelligence}, pages 671--708. Elsevier, 2008.

\bibitem{papageorgiou1991alinea}
M.~Papageorgiou, H.~Hadj-Salem, and J.-M. Blosseville.
\newblock Alinea: A local feedback control law for on-ramp metering.
\newblock {\em Transportation Research Record}, (1320):58--64, 1991.

\bibitem{papamichail2010coordinated}
I.~Papamichail, A.~Kotsialos, I.~Margonis, and M.~Papageorgiou.
\newblock Coordinated ramp metering for freeway networks--a model-predictive
  hierarchical control approach.
\newblock {\em Transportation Research Part C: Emerging Technologies},
  18(3):311--331, 2010.

\bibitem{papamichail2010heuristic}
I.~Papamichail, M.~Papageorgiou, V.~Vong, and J.~Gaffney.
\newblock Heuristic ramp-metering coordination strategy implemented at monash
  freeway, australia.
\newblock {\em Transportation Research Record: Journal of the Transportation
  Research Board}, 2178(1):10--20, 2010.

\bibitem{pisarski:hal-00727783}
D.~Pisarski and C.~Canudas~de Wit.
\newblock {Optimal Balancing of Road Traffic Density Distributions for the Cell
  Transmission Model}.
\newblock In {\em {51st IEEE Conference on Decision and Control (CDC 2012)}},
  Maui, Hawaii, United States, December 2012.

\bibitem{rasmussen2004gaussian}
C.E. Rasmussen.
\newblock Gaussian processes in machine learning.
\newblock In {\em Advanced lectures on machine learning}, pages 63--71.
  Springer, 2004.

\bibitem{richards1992RP}
B.L. Richards and R.J. Mooney.
\newblock Learning relations by pathfinding.
\newblock In {\em Proceedings of the Tenth National Conference on Artificial
  Intelligence}, AAAI'92, pages 50--55. AAAI Press, 1992.

\bibitem{domingos2006markov}
M.~Richardson and P.M. Domingos.
\newblock Markov logic networks.
\newblock {\em Machine Learning}, 62(1-2):107--136, 2006.

\bibitem{schmitt2016sufficient}
M.~Schmitt, C.~Ramesh, and J.~Lygeros.
\newblock Sufficient optimality conditions for distributed, non-predictive ramp
  metering in the monotonic cell transmission model.
\newblock submitted to Journal of Transportation Research, Part B:
  Methodological., 2016.

\bibitem{anskarl-TOCL15}
A.~Skarlatidis, G.~Paliouras, A.~Artikis, and G.A. Vouros.
\newblock {Probabilistic Event Calculus for Event Recognition}.
\newblock {\em ACM Transactions on Computational Logic}, 16(2):11:1--11:37,
  February 2015.

\bibitem{Starkeetal15}
S.~Starke, N.~Cooke, A.~Howes, N.~Morar, and C.~Baber.
\newblock {Visual sampling in a road traffic control management control room}.
\newblock In {\em {International Conference on Ergonomics and Human Factors}},
  pages 503--511. Taylor and Francis, 2015.

\bibitem{ThomasCook2005}
J.J. Thomas and K.A. Cook.
\newblock {\em Illuminiating the Path: the research and development agenda for
  visual analytics,}.
\newblock National Visualization and Analytics Center, Pacific Northwest
  National Laboratory, Richland, WA, 2005.

\bibitem{Vicente99}
K.J. Vicente.
\newblock {\em Cognitive work analysis: Toward safe, productive, and healthy
  computer-based work.}
\newblock Lawrence Erlbaum Associates, New York, 1999.

\bibitem{Wasserkrug2012}
S.~Wasserkrug, A.~Gal, O.~Etzion, and Y.~Turchin.
\newblock {Efficient processing of uncertain events in rule-based systems}.
\newblock {\em {IEEE Transactions on Knowledge and Data Engineering}},
  24(1):45--58, 2012.

\end{thebibliography}

\end{document}